\documentclass[runningheads]{llncs}

\usepackage{eccv}

\usepackage{eccvabbrv}

\usepackage{graphicx}
\usepackage{booktabs}
\usepackage{multirow}
\usepackage{booktabs}
\usepackage{algorithm}  
\usepackage{algpseudocode}  
\usepackage{amsmath}  
\usepackage[accsupp]{axessibility}  

\usepackage{hyperref}

\usepackage{orcidlink}
\usepackage{marvosym}
\usepackage{pdfpages}

\begin{document}

\title{Few-shot Defect Image Generation based on Consistency Modeling} 


\author{Qingfeng Shi\inst{1,2} \orcidlink{0009-0000-8480-9314}\and
Jing Wei\inst{1,2} \orcidlink{0000-0001-7697-8070}\and
Fei Shen\inst{1,2,3}\textsuperscript{(\Letter)}\orcidlink{0000-0001-9263-4489} \and
Zhengtao Zhang\inst{1,2,3}\orcidlink{0000-0003-1659-7879}}

\authorrunning{Q.~Shi et al.}

\institute{Institute of Automation, Chinese Academy of Sciences \and
School of Artificial Intelligence, University of Chinese Academy of Sciences \and
CASI Vision Technology CO., LTD., Luoyang, China\\
\email{\{shiqingfeng2022, weijing2020, fei.shen, zhengtao.zhang\}@ia.ac.cn}}

\maketitle
\begin{abstract}
Image generation can solve insufficient labeled data issues in defect detection. Most defect generation methods are only trained on a single product without considering the consistencies among multiple products, leading to poor quality and diversity of generated results. To address these issues, we propose DefectDiffu, a novel text-guided diffusion method to model both intra-product background consistency and inter-product defect consistency across multiple products and modulate the consistency perturbation directions to control product type and defect strength, achieving diversified defect image generation. Firstly, we leverage a text encoder to separately provide consistency prompts for background, defect, and fusion parts of the disentangled integrated architecture, thereby disentangling defects and normal backgrounds. Secondly, we propose the double-free strategy to generate defect images through two-stage perturbation of consistency direction, thereby controlling product type and defect strength by adjusting the perturbation scale. Besides, DefectDiffu can generate defect mask annotations utilizing cross-attention maps from the defect part. Finally, to improve the generation quality of small defects and masks, we propose the adaptive attention-enhance loss to increase the attention to defects. Experimental results demonstrate that DefectDiffu surpasses state-of-the-art methods in terms of generation quality and diversity, thus effectively improving downstream defection performance. Moreover, defect perturbation directions can be transferred among various products to achieve zero-shot defect generation, which is highly beneficial for addressing insufficient data issues. The code are available at https://github.com/FFDD-diffusion/DefectDiffu.
\end{abstract}
\section{Introduction}
\label{sec:intro}
Detection algorithms based on deep learning are crucial to industrial applications such as manufacturing\cite{Singh_Desai_2023}, transportation\cite{Liu_Liu_Zheng_Wang_Mao_Qiu_Ling_2023}, and power systems \cite{Ahmed_M_R_Mathur_2020}. Efficient defect detection methods facilitate the identification of anomalies from normal products\cite{schluter2022natural}. However, the performance of detection algorithms is often positively correlated with the number of annotated samples. With the improvement of production processes, defects become rarer in production lines and occupy only a small portion of the object\cite{Zavrtanik_Kristan_Skočaj}, posing challenges in data collection and annotation. Recently, image synthesis utilizing Generative Adversarial Networks (GANs)\cite{10.5555/2969033.2969125} and Denoising Diffusion Probabilistic Model (DDPM)\cite{Ho_Jain_Abbeel_2020} has attracted a surge of research\cite{kulikov2023sinddm}, leading to impressive progress in generating photo-realistic images\cite{tang2020xinggan}\cite{crowson2022vqgan}\cite{KowalskiECCV2020}. Recent works\cite{Zhang_Cui_Hung_Lu_2021}\cite{Niu_Li_Wang_Lin_2020}\cite{Niu_Li_Wang_Peng_2022}\cite{Niu_Peng_Li_Wang}\cite{Duan_Hong_Niu_Zhang_2023}\cite{Hu_Zhang_Yi_Du_Chen_Liu_Wang_Wang_2023} have demonstrated the feasibility of generating photo-realistic defect images to address the scarcity of defect images. However, existing methods face challenges when dealing with limited defect training samples, such as overfitting, poor diversity, and inaccurate annotations. 

\begin{figure}[tb]
  \centering
  \includegraphics[height=4.5cm]{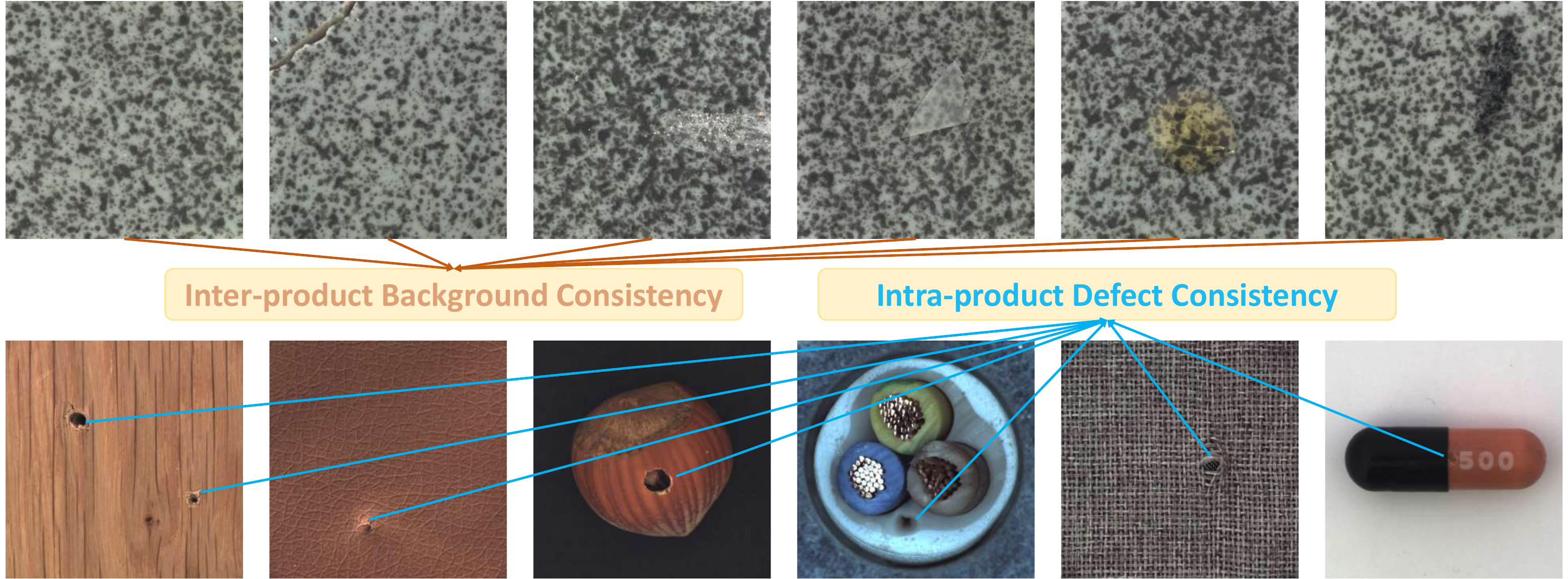}
  \caption{Example of intra-product background and inter-product defect consistency}
  \label{fig1}
\end{figure}

These challenges stem from the fact that current methods primarily focus on tailored designs for specific scenarios, neglecting to address two key consistencies in industrial defect images. Firstly, various defects within a single product all originate from normal regions. As depicted in the first row of \cref{fig1}, normal regions of defect images and normal images are consistent, highlighting intra-product background consistency. Secondly, as shown in the second row of \cref{fig1}, defects of the same category across various products exhibit morphological similarities, illustrating inter-product defect consistency. Leveraging these consistencies, defect image generation can be divided into two steps: normal feature generation and defect feature transformation. Notably, the generation direction for the normal background should remain consistent within the same product, while the defect generation direction should be uniform across similar defects of different products, with the direction scale reflecting the strength of the defect. Compared with modeling the defect distribution of a single type of product, modeling the two consistencies across multiple products enriches available training data for generation. Therefore, modeling the two consistencies is crucial for generating high-quality and diversified defect images with limited training samples.
Recently, the pre-trained vision-language model CLIP\cite{radford2021learning} has showcased remarkable zero-shot capability, displaying robust generalization. This prowess is largely attributed to that CLIP aligns text and images to effectively model the consistency of target objects. In downstream tasks, CLIP is often utilized to model consistency information among images. In anomaly detection, WinCLIP\cite{Jeong_Zou_Kim_Zhang_Ravichandran_Dabeer_2023} and AnomalyCLIP\cite{Zhou_Pang_Tian_He_Chen_2023} leverage the pre-trained CLIP to model consistency in anomalies, achieving zero-shot anomaly detection. In image generation, text-to-image generation based on diffusion models has shown promising results. Text prompts are encoded by the text encoder of CLIP to provide consistent conditions for generation and guide diffusion models to generate images conforming to the text prompts, enabling diverse generation. DreamBooth\cite{Ruiz_Li_Jampani_Pritch_Rubinstein_Aberman_2022} binds a unique identifier with the specific generation subject to achieve subject-driven generation. Custom Diffusion\cite{kumari2023multi} optimizes only a few parameters in the text-to-image conditioning mechanism by four images, enabling the model to learn new concepts and achieve combinations of multiple concepts.
 
Based on the aforementioned analysis, we propose a novel text-guided diffusion method DefectDiffu, to model the intra-product background consistency and inter-product defect consistency across multiple products, achieving controllable diversified defect image generation with few-shot training samples. Firstly, we introduce a disentangled integrated architecture, comprising three key parts: background, defect, and fusion. Then multiple single-object text conditions are inputted into each part through a text encoder to facilitate the disentanglement and integration of background and defect, achieving the modeling of the two consistencies. Additionally, we extract cross-attention maps from the defect part to derive accurate binary mask annotations. Secondly, we propose the double-free strategy to concretize the two consistency directions. Double-free allows for flexible control over product types and defect strengths by adjusting the consistency perturbation directions. Finally, addressing the issues of poor generation of small defects and inaccurate masks, we introduce the adaptive attention-enhanced loss to boost the attention on defect regions, improving the generation quality of small defects and masks. Furthermore, we explore the ability of DefectDiffu for zero-shot defect generation by transferring defect perturbation directions. The main contributions of this paper are as follows:\\
\noindent
1) We propose DefectDiffu to model intra-product background consistency and inter-product defect consistency across multiple products using text-guided disentangled integrated architecture, achieving controllable defect image generation with limited samples and obtaining masks during the generation.\\
\noindent
2) We propose the double-free strategy to concretize background and defect perturbation directions, allowing for adjustment of perturbation scales to control product types and defect strength, and further exploring zero-shot defect generation capabilities.\\
\noindent
3) We introduce the adaptive attention-enhanced loss to improve the generation quality of small defects and the accuracy of masks.
\section{Related Work}
\subsection{Controllable Image Generation}
\subsubsection{Diffusion-based generation.}
Diffusion-based generation. In recent years, the diffusion models have experienced rapid development in image generation due to their high generation quality and training stability, surpassing GANs and VAEs (Variational Autoencoders)\cite{Kingma2014} in terms of the realism and diversity of generated images, becoming the most prominent image generation model, particularly in controllable image generation. \cite{dhariwal2021diffusion} focuses on pre-training a classifier on noisy images, using the classification gradients of noisy images to guide category-controllable generation. \cite{ho2022classifier} proposes classifier-free guidance, simultaneously training both conditional and unconditional models, using their differences as a substitute for an explicit classifier. DiT\cite{peebles2023scalable} replaces the commonly used UNet architecture in diffusion models with transformers to learn parameters of adaptive normalization layers from conditions, achieving state-of-the-art performance on the ImageNet.
\subsubsection{Text-guided Image Generation.}
Recently, large-scale models based on diffusion models, such as DALLE 2\cite{Ramesh_Dhariwal_Nichol_Chu_Chen}, Imagen\cite{Zhang_Agrawala}, and Stable diffusion\cite{Rombach_Blattmann_Lorenz_Esser_Ommer_2022}, have demonstrated impressive text-guided image generation capabilities, enabling precise control over the generated semantic contents\cite{Zheng_2023_CVPR}\cite{zhao2023uni}. Cross-attention mechanisms are the most commonly used method for introducing textual guidance into generation models, where cross-attention maps reflect positions corresponding to textual tokens in generated images. To ensure the alignment between generated images and text prompts, Attend-and-Excite\cite{Chefer_Alaluf_Vinker_Wolf_Cohen-Or_2023}, MaskDiffusion\cite{zhou2023maskdiffusion}, and StructureDiffusion\cite{feng2022training} modify cross-attention to enhance the focus of models on each token. Some works achieve label generation and image editing by processing cross-attention maps. DiffuMask\cite{Wu_Zhao_Shou_Zhou_Shen_2023} extracts cross-attention maps and obtains pseudo-masks for generated images through affinity networks. Attn2Mask\cite{Yoshihashi_Otsuka_Doi_Tanaka_2023} uses post-processed cross-attention maps as pseudo-masks for training segmentation models. Prompt-to-Prompt dynamically changes cross-attention maps during the generation to enable flexible image editing. Text-guided models demonstrate the ability to learn high-level semantic information from images \cite{Ruiz_Li_Jampani_Pritch_Rubinstein_Aberman_2022} \cite{kumari2023multi} \cite{gal2022image}, allowing for customization generation by fine-tuning pre-trained models with a few images and textual descriptions.
\subsection{Defect Image Generation}
Defect image generation is widely used to address insufficient data issues in detection tasks. Most current methods employ GANs to generate defect images with binary masks  (e.g., DefectGAN\cite{zhang2021defect}, SDGAN\cite{Niu_Li_Wang_Lin_2020}, AnomalyGAN\cite{Liu_Liu_Zheng_Wang_Mao_Qiu_Ling_2023}). MDGAN\cite{machines10121239} synthesizes defects based on masks constructed from Perlin noise and defect annotations on real normal samples. \cite{Niu_Li_Wang_Peng_2022} and \cite{Niu_Li_Wang_Lin_2020} encode images into feature space to establish transformation directions between normal and defect images, and adjust transformation factors to achieve strength controllable defect image generation.  DFMGAN\cite{Duan_Hong_Niu_Zhang_2023} pre-trains StyleGAN v2\cite{9156570} on normal samples and fine-tunes with a few defect samples to update the proposed mask and defect generation modules. Currently, some works have explored few-shot defect image generation based on diffusion models. To improve generation quality, Defect-Gen\cite{yang2023defect} trains two diffusion models, the large model to determine the generated target structure and the small model for detail generation. However, these methods require training separate models for each product. Anomalydiffusion\cite{Hu_Zhang_Yi_Du_Chen_Liu_Wang_Wang_2023} generates masks by text flipping, adopting normal samples and masks as inputs to generate defect images for multiple products. However, predefined masks and normal samples limit the diversity of defect shapes and normal backgrounds. Moreover, the aforementioned methods fail to model the relationships between similar defects of multiple products, resulting in poor generalization. In contrast, DefectDiffu models image consistencies across multiple products and obtains masks based on cross-attention maps, thereby integrating features from multiple products to generate diversified defect images and avoid restrictions from predefined masks.
\section{Method}
\begin{figure}[tb]
  \centering
  \includegraphics[height=5cm]{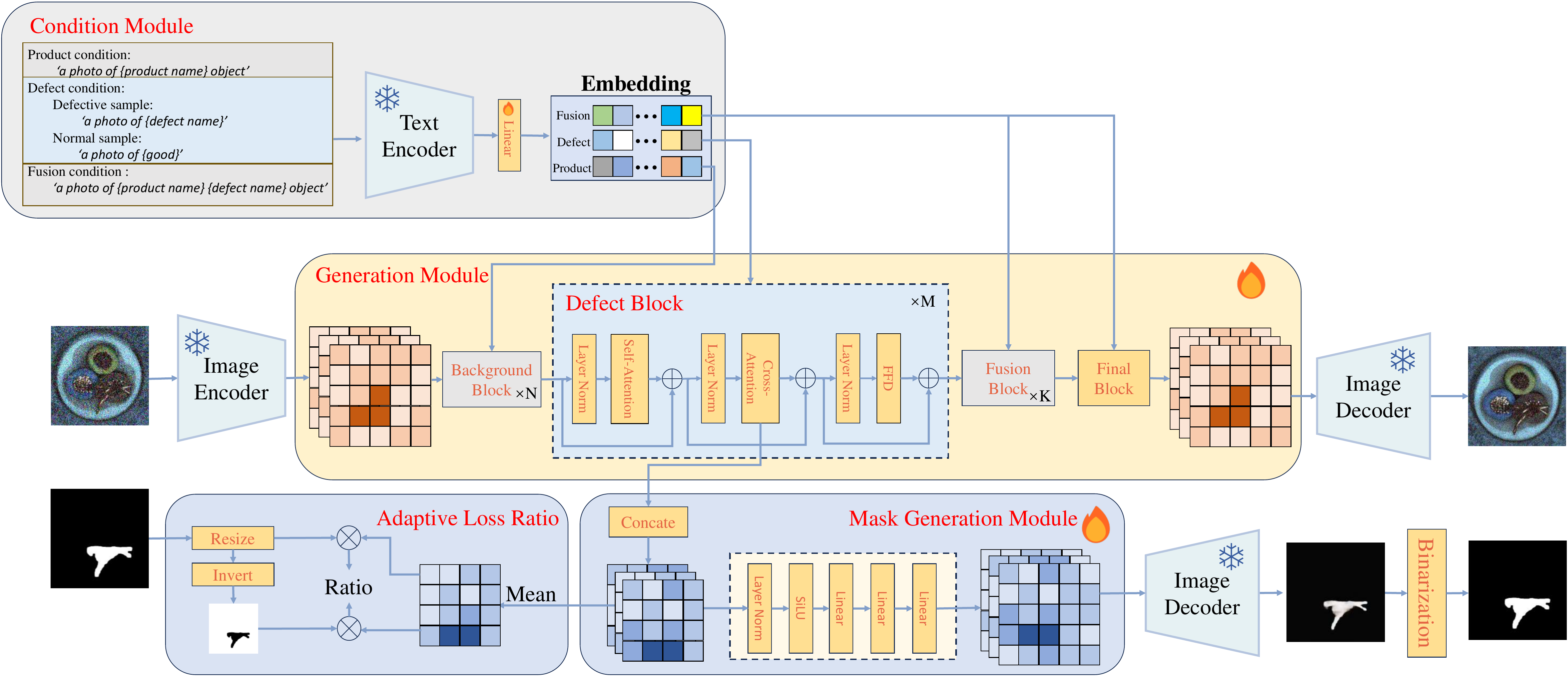}
  \caption{Overall framework of DefectDiffu. During training, real images and masks are encoded by \textit{Image Encoder} into feature spaces, which are then used for training the \textit{Generation Module} and the \textit{Mask Generation Module}. The real masks are resized for calculating the \textit{Adaptive Loss Ratio}. During testing, generated image features and mask features are decoded by \textit{Image Decoder} to obtain image-mask pairs.}
  \label{fig2}
\end{figure}

As illustrated in \cref{fig2}, DefectDiffu adopts pre-trained DiT as backbones, consisting of the Condition Module, Generation Module, and the Mask Generation Module. The condition module employs the CLIP Text Encoder to introduce consistent text prompts into corresponding parts of the Generation Module. The generation module adopts a disentangled integrated architecture to disentangle the generation of normal backgrounds and defects, and then integrate them to generate realistic images under the guidance of text prompts. The mask generation module integrates cross-attention maps of the Defect Blocks to obtain mask annotations corresponding to generated images. Furthermore, the Adaptive Loss Ratio is designed to adjust the loss weights of the defect regions, enhancing the attention of DefectDiffu to defects. Next, we propose a double-free strategy to concretize the modeled consistencies, thereby achieving controllable generation of the product type and defect strength by modulating denoising results.
\subsection{Preliminaries}
Image generation based on DDPM consists of the forward noise addition process and the reverse denoising process. In the forward process, Gaussian noise is added into the input image $x_0$ for $T$ times to obtain the pure noisy image $x_T$,
\begin{equation}
  q(x_{1:T}\vert x_0):=\prod \limits_{t=1}^T q(x_t\vert x_{t-1}),q(x_t\vert x_{t-1}):=N(x_t;\sqrt{1-\beta_t}x_{t-1},\beta_t I)
  \label{eq:important}
\end{equation}
where  $\beta_t$ is the variance at timestep $t$. In the reverse denoising process, the noise prediction model  $\epsilon_{\theta}$ recovers $x_0$ from the noisy image $x_t$, where $\theta$ represents the learnable model parameters. $\epsilon_{\theta}$  takes  $x_t$ and the current time step $t$ as inputs to predict the removed noise at this time step. Furthermore, when implementing controllable image generation with text, images, or other information as condition $c$, $\epsilon_{\theta}$  takes $x_t$, the time step $t$, and the condition $c$ as inputs to predict the noise. The training objective of the model is:
\begin{equation}
  \min_{\theta}L=\mathbb E_{x_t ,\epsilon}\left[\Vert \epsilon-\epsilon_{\theta}(x_t,t,c)\Vert_2^2\right]
  \label{eq:object}
\end{equation}
\subsection{Text-guided Disentangled Integrated Architecture}
To model the intra-product background consistency and inter-product defect consistency, it is necessary to provide conditions that include consistency information for DefectDiffu and design architecture to achieve disentangled generation guided by consistency information. 
Firstly, as shown in \cref{fig2}, we adopt product names and defect names as intra-product consistency and inter-product defect consistency text prompts respectively, employing the CLIP text encoder to construct consistent guidance. Previous text-to-image generation methods used \textbf{\emph{A photo of \{object 1\} \{object 2\}  \{object n\}}} to control the generation of multiple objects, which is not suitable for our work. The reason is that the text encoder encodes multiple objects in one text prompt into a fused output. Although this fused output can guide the model to generate high-quality images, it is difficult to disentangle multiple objects in the generation and may cause catastrophic neglect \cite{Chefer_Alaluf_Vinker_Wolf_Cohen-Or_2023}. However, disentangling defects and backgrounds is the key point in modeling two consistencies of multiple products. Therefore, we use two single-object prompts as conditions to model consistencies, achieving the disentangled generation of backgrounds and defects. In addition, compared to the UNet-based diffusion architecture, DiT has a scalable hierarchical stacking architecture based on transformers. It allows us to introduce different conditions into each part of the model to learn the disentanglement generation of multiple objects.  

Therefore, as shown in \cref{fig2}, combining the advantages of single-object guidance and the flexibility of DiT, we propose the Text-guided Disentangled Integrated Architecture (TDIA). TDIA divides DiT into three parts, background, defect, and fusion. Background text embedding for modeling background consistency is introduced into the background part, defect text embedding for modeling defect consistency is introduced into the defect part, and the fusion part adopts a defect-background fusion text embedding to generate realistic defect images. Designed text prompt templates for each part are as follows:

\noindent 
\hspace{0.5cm}Consistency condition $c_p$ for Background Block:
\begin{center}
\textbf{\emph{A photo of \{product name\}}}
\end{center}
\hspace{0.5cm}Consistency condition $c_d$ for Defect Block:
\begin{center}
\textbf{\emph{A photo of \{defect name\}}}
\end{center}
\hspace{0.5cm}Fusion condition $c_f$ for Fusion Block:
\begin{center}
\textbf{\emph{A photo of \{defect name\} \{product name\}}}
\end{center}
\subsection{Double-free}
Although DefectDiffu can generate diversified defect image $\epsilon_{\theta}(x_t,t,c_d,c_p)$  conditioned on $c_p$ and  $c_d$ after modeling two consistency by TDIA, the strength of generated defects can not be controlled and is limited to the training set distribution, which affects the diversity of generation. We hypothesize that there should be a perturbation direction along which normal images gradually become defect images, and the perturbation scale reflects the defect strength. Therefore, we perceive controllable defect image generation as a two-stage perturbation process involving both background and defect, with the defect strength adjustable by varying the perturbation scale. We set ${F_p^t}$  and  ${F_d^t}$ as perturbation directions concerning two consistent conditions $c_p$ and $c_d$ at time step $t$. Then controllable generation can be achieved by adjusting the perturbation scale $w_d$ and $w_p$ to update the denoised results
\begin{equation}
  \epsilon_{\theta}(x_t,t,c_d,c_p)=\epsilon_{\theta}(x_t,t,\varnothing,\varnothing)+w_p{F_d^t}+w_d{F_p^t}
  \label{eq:3}
\end{equation}
In order to calculate the two perturbation directions, we propose the double-free strategy based on the classifier-free guidance, achieving background and defect strength controllable generation beyond the diversity of the training set. First, considering $\epsilon_{\theta}(x_t,t,c_g,c_p)$  as conditional generation of normal background, where $c_g$ is \textbf{\emph{good}}, and unconditional background generation is realized by setting $c_p$ to null to get $ \epsilon_{\theta}(x_t,t,c_g,\varnothing)$, and the difference  ${F_p^t} = \epsilon_{\theta}(x_t,t,c_g,c_p)-\epsilon_{\theta}(x_t,t,c_g,\varnothing)$ is the perturbation direction for intra-product consistency, realizing controllable background generation. On this basis, inter-product defect consistency guidance is regarded as conditional generation of defect $\epsilon_{\theta}(x_t,t,c_d,\varnothing)$, corresponding unconditional generation is $\epsilon_{\theta}(x_t,t,c_g,\varnothing)$, and the difference  ${F_d^t} = \epsilon_{\theta}(x_t,t,c_d,\varnothing)-\epsilon_{\theta}(x_t,t,c_g,\varnothing)$ is perturbation direction for inter-product defect consistency. Then we set the perturbation direction scale $w_d$ of defects relative to normal images to achieve the controllable generation of defect strength. Finally, the \cref{eq:3} can be simplified as:
\begin{align}
    \epsilon_{\theta}(x_t,t,c_d,c_p)& =\epsilon_{\theta}(x_t,t,c_g,\varnothing)+w_p(\epsilon_{\theta}(x_t,t,c_g,c_p)-\epsilon_{\theta}(x_t,t,c_g,\varnothing)) \; \\
  & +w_d(\epsilon_{\theta}(x_t,t,c_d,\varnothing)-\epsilon_{\theta}(x_t,t,c_g,\varnothing))\nonumber
\end{align}
where $w_p$ and $w_d$ control the representation of the background and the strength of the defect respectively. The algorithm of double-free is shown in \cref{alg:Framwork}.
\begin{algorithm}[t]  
  \caption{ Double-free guidance.}  
  \label{alg:Framwork}  
  \begin{algorithmic}[1]
    \Require  
      Training data $(x,c_d^r,c_p^r)$ from the dataset, $p_1$: probability of unconditional product, $p_2$: probability of unconditional defect, $c_d=c_d^r,c_p=c_p^r$
    \Ensure  
      Condition modified data, $(x,c_d,c_p)$;  
    \State with probability $p_1$, $c_p\rightarrow \varnothing$\\
    \textbf{If}   {$c_d^r$ is not \textbf{\emph{good}}}:\\ 
    \qquad with probability $p_2$, $c_d\rightarrow$\textbf{\emph{good}};\\
    \qquad with probability $p_1$, $c_p\rightarrow\varnothing$;
    \label{code:fram:select} \\  
    \Return $(x,c_d,c_p)$;  
    \end{algorithmic}
\end{algorithm}  

\subsection{Mask Generation}
When training segmentation models with generated images, obtaining pixel-level mask annotations manually is laborious and time-consuming. Therefore, as shown in \cref{fig2}, we employ cross-attention mechanisms into the defect blocks and combine them with the mask generation module to simultaneously generate corresponding masks while generating defect images. Cross-attention is applicable for fusing the embeddings of the visual and textual features and producing spatial attention maps for each textual token. The noisy feature  $\varphi(x_t)\in\mathbb R^{H\times W\times C}$ is projected to query vector $Q\in\mathbb R^{H\times W\times d}$, where $H$, $W$, and $C$ is the length, width, number of channels of the tensor respectively, and $d$ is the dimension of $Q$. The text embedding $p\in\mathbb R^{l\times C}$ with length $l$ is projected to get key vector $K\in\mathbb R^{l\times d}$ and value vector $V\in\mathbb R^{l\times d}$. Then cross-attention is calculated as,
\begin{equation}
    CrossAttention=Softmax(\frac{QK^T}{\sqrt{d}})
  \label{eq:important}
\end{equation}
Next, we concatenate the cross-attention maps of multiple defect blocks along the channel dimension, followed by the mask generation module to obtain mask features $M_{f}$, and then decode  $M_{f}$ to obtain pixel-level masks $M_{pre}$  with values between [0, 255] to locate generated defect positions. The real mask is encoded to feather $Mask_f$ and the mask prediction loss is:
\begin{equation}
    loss_m=\Vert M_{f}-Mask_f\Vert_2^2
  \label{eq:important}
\end{equation}
With increasing time steps, the high-level semantic structure of the image will gradually emerge. Therefore, for accurate defect localization, we compute the average mask $M_{pre}=\frac{1}{100}\Sigma_{t=0}^{99}M_{pre}^t$  obtained from the last 100 time steps during testing and use an iterative method to binarize $M_{pre}$ to obtain the final mask.
\subsection{Adaptive Attention-enhanced Loss}
During training, we found that DefectDiffu tends to overlook defect features and generate normal images easily. Additionally, the generated masks are inaccurate for cases involving small defects or multiple defects within a single image. The reason lies in the fact that compared to normal regions, the features of defect regions with small proportions are more difficult to capture. Moreover, as shown in \cref{eq:object}, the training objective of the diffusion model is to minimize the global reconstruction loss, where the reconstruction loss of defects contributes less to the overall loss. Consequently, the model has insufficient attention to defect regions, leading to poor reconstruction effects and inaccurate annotations. 
To address these issues, we propose the adaptive attention-enhanced loss based on cross-attention maps to increase the loss weight of defect regions, thereby enhancing the attention of DefectDiffu on defects. Firstly, we calculate the average cross-attention $ Map$ from extracted cross-attention maps of the defect block:
\begin{equation}
    Map=\frac{1}{M}\sum_{i=0}^{M} Softmax(\frac{Q_iK^T}{\sqrt{d}})
  \label{eq:important}
\end{equation}
where $M $ is the number of extracted cross-attention maps. The real $ Mask$ is scaled to $mask$ with the same size of $Map$ to extract the attention for defect regions and normal regions, and then calculate the sum of each to obtain the adaptive attention ratio: 
\begin{equation}
    Rat=\frac{ \sum_{i,j}(1-mask)_{i,j}\cdot Map_{i,j} }{ \sum_{i,j}mask_{i,j}\cdot Map_{i,j} +\alpha }
  \label{eq:ratio}
\end{equation}
where $i$ and $j$ are height and width of $Map$, $\alpha$ is a constant used to prevent division by zero. The lower the attention allocated to the defect regions, the higher the $Rat$. 
To prevent the imbalance between defect and background regions caused by excessively high weights on defect regions, we further modify the magnitude of the attention ratio $Rat$ to get $R$,
\begin{equation}
    R(Rat)=\left\{
                \begin{array}{ll}
                  8,& Rat\geq 8\\
                  Rat,& 2<Rat<8\\
                  2,& Rat\leq 2
                \end{array}
                \label{rat}
              \right.
\end{equation}
then the improved defect region loss is:
\begin{equation}
    loss_d=R\Vert mask\cdot\epsilon_{\theta}(z_t,t,c_d,c_p)-mask\cdot\epsilon\Vert_2^2
  \label{eq:important}
\end{equation}
Finally, the training objective of DefectDiffu is:
\begin{equation}
    \min_{\theta} loss=\Vert \epsilon_{\theta}(z_t,t,c_d,c_p)-\epsilon\Vert_2^2 + loss_d+\lambda loss_m
  \label{eq:loss}
\end{equation}
 where $ \lambda$ is hyperparameter.

\subsection{Zero-shot Generation}
\label{eq:zero-shot method}
Based on the TDIA and double-free strategy, DefectDiffu can transfer defect features across normal backgrounds of different products, enabling zero-shot generation of defects. Only normal samples of the target product $T$ are used in training, during testing, the perturbation direction of defect $c_d^s$ learned from source products is added into the generation of normal images of $T$, resulting in generating defect samples of $T$ with defect $c_d^s$. Specifically, during testing, the model is simultaneously fed with product prompt $c_T$ and defect prompt $c_d^s$. Then the  perturbation direction of $c_d^s$ modeled on other products is computed to update the denoised results at time step $t$, 
\begin{align}
    \epsilon_{\theta}(x_t,t,c_d^s,c_T)& =\epsilon_{\theta}(x_t,t,c_g,\varnothing)+w_p(\epsilon_{\theta}(x_t,t,c_g,c_T)-\epsilon_{\theta}(x_t,t,c_g,\varnothing)) \; \\
  & +w_d(\epsilon_{\theta}(x_t,t,c_d^s,\varnothing)-\epsilon_{\theta}(x_t,t,c_g,\varnothing))\nonumber
\end{align}
 
Hence, DefectDiffu can generate defect images of $T$ without requiring corresponding real defect images for training, which is attributed to that we successfully model the perturbation direction of the inter-product defect consistency.  
\begin{figure}[tb]
  \centering
  \includegraphics[height=8cm, width=\linewidth]{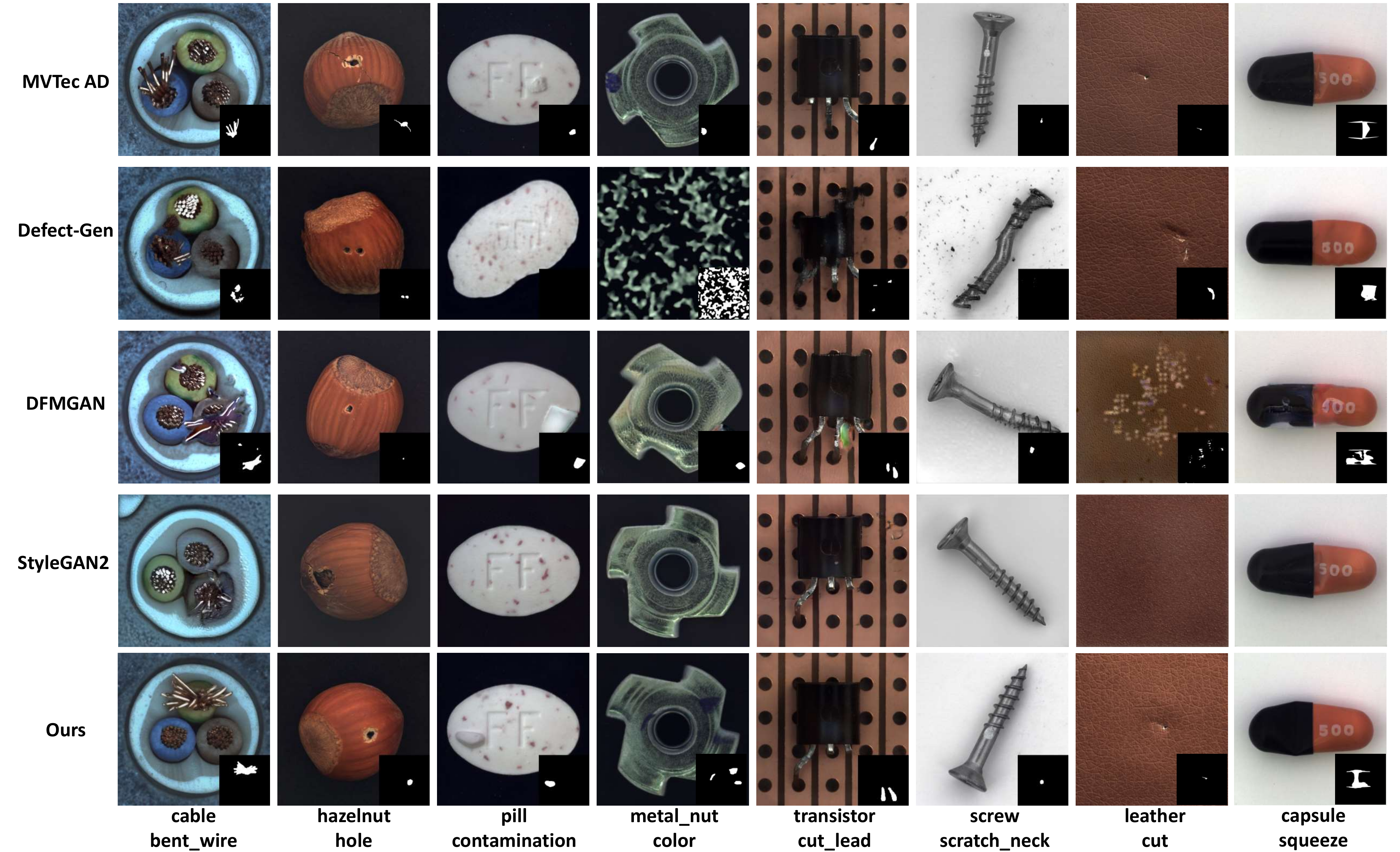}
  \caption{Comparison of the generation results.}
  \label{fig3}
\end{figure}
\section{Experiment}
\subsection{Experiment Settings}
\subsubsection{Dataset and Evaluation Metrics.}
 We conduct experiments on the MVTecAD dataset to validate the effectiveness of DefectDiffu. We select half of the images from the MVTecAD-test as the training set for image generation and baseline in segmentation experiments, with the remaining images used as the test set to evaluate the effectiveness of generated images in defect detection. All images are resized to 512×512. FID (Fréchet Inception Distance) and LPIPS (Learned Perceptual Image Patch Similarity) are calculated for quality and diversity evaluation respectively. A lower FID indicates higher generation quality. A higher LPIPS indicates better diversity. For the downstream application, we calculated mIOU (mean Intersection over Union), F1 score, mAP (mean Average Precision), and AUROC (Area Under the Receiver Operating Characteristic Curve) to evaluate the effectiveness of generated image-mask pairs, higher values indicate better performance.
\subsubsection{Implementation.}We set the N, M, and K in \cref{fig2} to 10, 10, and 8 respectively. The parameters of the corresponding modules are initialized with pre-trained DiT in ImageNet with the resolution of $512\times 512$. We set $\lambda =0.2$ in \cref{eq:loss}, $p_1= 0.2$, $p_2=0.2$, $\alpha=0.001$ in \cref{eq:ratio}. We use an iterative thresholding method for the binarization of generated masks. See the appendix for more details. 
\subsubsection{Comparison Methods.} We employ StyleGAN2, DFMGAN, and Defect-Gen as comparison methods. StyleGAN2 has shown promising results in natural image generation and is commonly used as a foundational framework in industrial image generation. DFMGAN focuses on few-shot defect generation based on StyleGAN2. Defect-Gen utilizes two diffusion models to generate defects and their corresponding masks. They are the current state-of-the-art methods in industrial image generation with annotated masks. It is worth noting that these models need to be trained separately for each product. See the appendix for implementation details of the comparison methods.
\begin{figure}[tb]
  \centering
  \includegraphics[height=8cm]{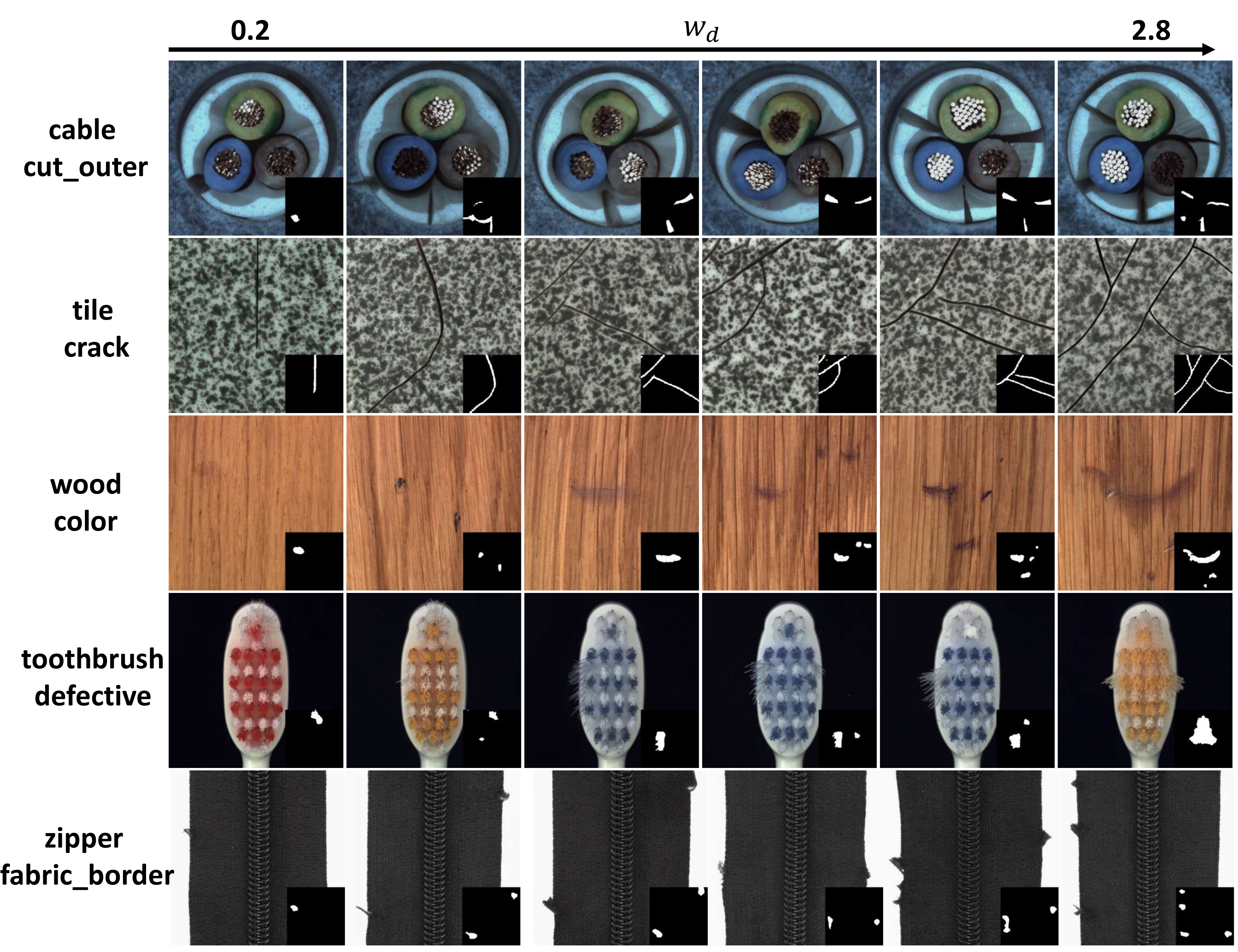}
  \caption{Generation results under different strengths, with defect strength gradually increasing from left to right.}
  \label{fig4}
\end{figure}
\begin{table}[t]
  \centering
  \caption{Evaluation results of quality and diversity for comparison and ablation experiments. Our DefectDiffu achieves optimal results on the MVTec dataset.}
  \resizebox{1\textwidth}{!}{
    \begin{tabular}{c|cc|cc|cc|cc|cc|cc|c|c}
    \hline
    \multicolumn{1}{c|}{\multirow{2}[2]{*}{Category}} & \multicolumn{2}{c|}{DefectDiffu} & \multicolumn{2}{c|}{Defect-Gen} & \multicolumn{2}{c|}{DFMGAN} & \multicolumn{2}{c|}{StyleGAN2} & \multicolumn{2}{c|}{w/o fusion} & \multicolumn{2}{c|}{w/o AAL} &OPT    & single \\
          & FID   & LPIPS & FID   & LPIPS & FID   & LPIPS & FID   & LPIPS & FID   & LPIPS & FID   & LPIPS & FID   & LPIPS \\
    \hline
    bottle & \textbf{73.48} & \textbf{0.16} & 124.86 & 0.10  & 130.69 & 0.13  & 112.79 & 0.08  & 132.20 & 0.08  & 134.76 & 0.08  & 127.12 & 0.09 \\
    cable & \textbf{101.69} & \textbf{0.29} & 158.82 & 0.24  & 111.38 & 0.27  & 144.33 & 0.17  & 112.48 & 0.10  & 108.30 & 0.10  & 185.85 & 0.28 \\
    capsule & 26.85 & \textbf{0.08} & 70.36 & 0.02  & 50.54 & 0.05  & \textbf{25.62} & 0.03  & 57.68 & 0.05  & 60.35 & 0.03  & 70.43 & 0.07 \\
    carpet & \textbf{40.31} & \textbf{0.17} & 51.35 & 0.13  & 45.09 & 0.16  & 40.34 & 0.14  & 44.56 & 0.10  & 43.85 & 0.10  & 77.25 & 0.16 \\
    grid  & \textbf{62.55} & \textbf{0.35} & 136.62 & 0.30  & 84.58 & 0.34  & 90.91 & 0.33  & 75.91 & 0.24  & 81.45 & 0.24  & 161.24 & \textbf{0.35} \\
    hazelnut & \textbf{70.18} & 0.17  & 434.55 & 0.13  & 378.17 & \textbf{0.20} & 435.11 & \textbf{0.20} & 420.76 & 0.15  & 464.62 & 0.13  & 185.90 & 0.17 \\
    leather & \textbf{67.19} & 0.15  & 470.15 & 0.10  & 95.84 & \textbf{0.31} & 369.94 & 0.14  & 96.94 & 0.10  & 88.45 & 0.09  & 435.32 & 0.14 \\
    metal\_nut & \textbf{59.46} & 0.29  & 168.21 & \textbf{0.47} & 151.68 & 0.33  & 87.56 & 0.18  & 158.56 & 0.25  & 172.42 & 0.19  & 232.26 & 0.29 \\
    pill  & \textbf{46.23} & 0.15  & 258.37 & \textbf{0.42} & 133.07 & 0.14  & 227.78 & 0.02  & 162.59 & 0.06  & 152.54 & 0.07  & 186.61 & 0.14 \\
    screw & \textbf{26.11} & 0.27  & 217.84 & \textbf{0.42} & 71.06 & 0.26  & 32.59 & 0.26  & 70.54 & 0.17  & 71.36 & 0.18  & 120.84 & 0.26 \\
    tile  & \textbf{120.39} & \textbf{0.36} & 333.24 & 0.24  & 162.14 & 0.18  & 136.40 & 0.26  & 152.25 & 0.22  & 158.48 & 0.22  & 162.18 & 0.34 \\
    toothbrush & 110.59 & 0.14  & 196.56 & 0.13  & 224.22 & \textbf{0.24} & \textbf{48.08} & 0.12  & 220.34 & 0.19  & 212.30 & 0.16  & 418.58 & 0.13 \\
    transistor & 69.25 & 0.24  & 173.35 & 0.21  & 142.69 & \textbf{0.28} & \textbf{67.07} & 0.26  & 127.27 & 0.19  & 123.21 & 0.11  & 204.25 & 0.20 \\
    wood  & \textbf{128.26} & 0.27  & 333.03 & 0.19  & 315.69 & 0.23  & 188.20 & \textbf{0.34} & 331.29 & 0.19  & 325.66 & 0.16  & 305.98 & 0.23 \\
    zipper & \textbf{37.74} & 0.12  & 378.24 & \textbf{0.48} & 80.97 & 0.07  & 78.18 & 0.16  & 89.36 & 0.05  & 93.80 & 0.05  & 89.48 & 0.11 \\
    \hline
    Average & \textbf{69.35} & 0.21  & 233.70 & \textbf{0.24} & 145.19 & 0.21  & 138.99 & 0.18  & 149.83 & 0.14  & 152.77 & 0.13  & 197.55 & 0.20 \\
    \hline
    \end{tabular}%
    }
  \label{tab:table1}%
\end{table}%
\subsection{Comparison of Generation Results}
\cref{fig3} shows the generated results of different methods, see the appendix for more results. It can be seen that Defect-Gen fails to model realistic high-level semantic features and the generated defects do not match real defect characteristics. DFMGAN generates unrealistic defects and contains numerous artifacts. StyleGAN2 fails to converge on leather and exhibits artifacts in products like cable. In contrast, DefectDiffu achieves good convergence across all 15 products. The generated results conform to real features, without mode collapse or artifacts, and the masks are accurate, validating the effectiveness of the proposed multi-product consistency modeling for few-shot image generation. Additionally, \cref{fig4} demonstrates the control of defect strength utilizing $w_d$. As $w_d$ increases, the area of generated defects expands, and the number of defect regions increases, indicating a gradual increase in the severity of product damage. 

\cref{tab:table1} shows that our method achieves the optimal generation quality and good diversity on the MVTec dataset. Although the generation quality of toothbrush is slightly inferior to StyleGAN2, DefectDiffu simultaneously generates defect masks, which is more beneficial for training segmentation methods. Despite DFMGAN and Defect-Gen achieving higher LPIPS on some products, this is due to the introduction of unrealistic artifacts and the loss of real semantic features. It can be seen in \cref{fig3} and \cref{tab:table1} that the generation quality of these products is poor, corresponding to high FID. Clearly, diversity without adherence to real features is meaningless. In contrast, DefectDiffu achieves optimal generation quality on most products while maintaining good diversity. 
\subsection{Comparison of Detection Results. }
\begin{table}[t]\small
  \centering
  \caption{Comparison on detection performance.}
    \resizebox{1\textwidth}{!}{
    \begin{tabular}{c|cccc|cccc|cccc|cccc}
    \hline
    \multicolumn{1}{c|}{\multirow{2}[2]{*}{Category}}  & \multicolumn{4}{c|}{Baseline} & \multicolumn{4}{c|}{DefectDiffu} & \multicolumn{4}{c|}{Defect-Gen} & \multicolumn{4}{c}{DFMGAN} \\
          & mIoU  & F1    & mAP   & AUC   & mIoU  & F1    & mAP   & AUC   & mIoU  & F1    & mAP   & AUC   & mIoU  & F1    & mAP   & AUC \\
  \hline
    bottle & 66.1  & 79.6  & 89.7  & 98.7  & 75.7  & 86.2  & \textbf{93.8} & \textbf{99.2} & \textbf{76.0} & \textbf{86.3} & 93.2  & 99.0  & 71.3  & 83.2  & 92.3  & 98.8 \\
    cable & 55.0  & 71.0  & 76.0  & 92.2  & \textbf{62.5} & \textbf{76.9} & \textbf{81.8} & \textbf{95.1} & 60.5  & 75.4  & 80.6  & \textbf{95.1}  & 58.4  & 73.7  & 78.0  & 91.8 \\
    capsule & 28.2  & 44.0  & 42.2  & 94.1  & \textbf{34.2} & \textbf{51.0} & 54.9  & \textbf{97.2} & 31.4  & 47.8  & 47.5  & 91.2  & 33.1  & 49.8  & \textbf{56.4}  & 96.8 \\
    carpet & 64.8  & 78.6  & 88.8  & \textbf{99.6}  & \textbf{69.8} & \textbf{82.2} & \textbf{90.9} & \textbf{99.6} & 67.4  & 80.5  & 89.8  & 99.3  & 66.1  & 79.6  & 88.8  & 99.5 \\
    grid  & 32.7  & 49.3  & 46.6  & 95.8  & \textbf{47.3} & \textbf{64.2} & \textbf{69.4} & \textbf{99.3} & 35.6  & 52.5  & 49.6  & 97.4  & 37.4  & 54.4  & 63.8  & 98.8 \\
    hazelnut & 85.0  & 91.9  & 96.9  & 99.6  & 87.2  & 93.2  & 97.5  & 99.7  & \textbf{87.4} & \textbf{93.3} & \textbf{97.9} & \textbf{99.8} & 83.1  & 90.8  & 96.4  & 99.5 \\
    leather & 67.4  & 80.5  & 89.6  & 99.8  & 67.9  & 80.9  & 88.1  & 99.7  & \textbf{68.3} & \textbf{81.1} & \textbf{89.9} & \textbf{99.8} & 65.3  & 79.0  & 88.4  & 99.8 \\
    metal\_nut & 86.9  & 93.0  & 97.6  & 99.3  & \textbf{91.5} & \textbf{95.6} & \textbf{99.0} & 99.7  & 89.5  & 94.5  & 98.8  & \textbf{99.8} & 73.0  & 84.4  & 92.4  & 97.9 \\
    pill  & 76.1  & 86.4  & 95.2  & 99.7  & \textbf{83.4} & \textbf{90.9} & \textbf{97.4} & \textbf{99.8} & 69.5  & 82.0  & 91.8  & 99.0  & 76.8  & 86.9  & 94.3  & 99.3 \\
    screw & 26.1  & 41.4  & 47.0  & 98.3  & \textbf{48.9} & \textbf{65.7} & \textbf{76.6} & \textbf{99.7} & 43.0  & 60.1  & 69.9  & 98.5  & 32.7  & 49.3  & 57.9  & 98.1 \\
    tile  & 86.7  & 92.9  & 98.2  & 99.8  & \textbf{89.7} & \textbf{94.6} & 98.8  & \textbf{99.9} & 89.1  & 94.2  & \textbf{98.9} & \textbf{99.9} & 87.7  & 93.4  & 98.6  & 99.8 \\
    toothbrush & 34.5  & 51.3  & 56.6  & 95.8  & 44.0  & 61.1  & 60.2  & 96.2  & \textbf{44.8} & \textbf{61.8} & \textbf{60.7} & \textbf{97.0} & 34.2  & 31.4  & 47.8  & 58.3 \\
    transistor & 58.2  & 73.6  & 78.9  & 96.0  & \textbf{64.3} & \textbf{78.3} & \textbf{79.2} & \textbf{97.1} & 52.8  & 69.1  & 84.6  & 96.7  & 48.7  & 65.5  & 74.7  & 94.4 \\
    wood  & 68.3  & 81.1  & 89.5  & 97.7  & \textbf{69.4} & \textbf{81.9} & \textbf{89.8} & \textbf{98.3} & 69.3  & \textbf{81.9}  & 89.5  & 98.0  & 66.5  & 79.9  & 88.1  & 97.8 \\
    zipper & 67.9  & 80.8  & 87.7  & 99.3  & 70.7  & \textbf{82.9} & \textbf{90.2} & \textbf{99.6} & 68.6  & 81.4  & 88.9  & 99.4  & \textbf{70.8} & \textbf{82.9} & 89.2  & \textbf{99.6} \\
  \hline
    Average & 60.3  & 73.0  & 78.7  & 97.7  & \textbf{67.1} & \textbf{79.0} & \textbf{84.5} & \textbf{98.7} & 63.5  & 76.1  & 82.1  & 98.0  & 60.3  & 72.3  & 80.5  & 95.4 \\
  \hline
    \end{tabular}%
    }
  \label{tab:table2}%
\end{table}%
To validate the effectiveness of the generated results in industrial detection, we employ the real training set to train a UNet as a baseline and generate synthetic samples twice the size of real images by those methods capable of generating masks to augment the real dataset. Segmentation results are shown in \cref{tab:table2}. Compared to other methods, DefectDiffu can improve the performance of all products and achieve optimal segmentation results on most products. More results are provided in the appendix. In summary, DefectDiffu can efficiently improve the performance of detection methods and have high application value.
\subsection{Zero-shot Generation}
\begin{figure}[tb]
  \centering
  \includegraphics[height=1.8cm]{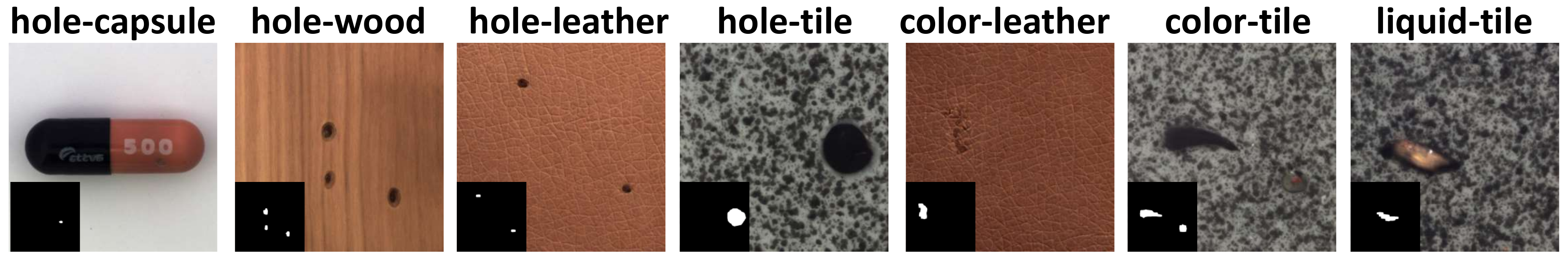}
  \caption{Zero-shot generation results. \textit{A-B} means to transfer source defect A to target product B.}
  \label{fig5}
\end{figure}
As outlined in \cref{eq:zero-shot method}, we validate the zero-shot generation capability of DefectDiffu, with results shown in \cref{fig5}. Although DefectDiffu only learns the features of normal images from the target product during training, the defect consistency perturbation directions, modeled from other products, effectively guide the generation of defects for the target product. This enables DefectDiffu to achieve zero-shot defect generation without the need for real defect images of the target product during training. See the appendix for more results.
\subsection{Ablation Experiments}
\subsubsection{Fusion part}. We remove the fusion part in the generation module, and the results are shown in \cref{tab:table1} w/o fusion. Without the fusion part, the semantic coherence between generated defects and backgrounds is disrupted, leading to a reduction in generation quality and diversity.
\subsubsection{AAL (Adaptive Attention-enhanced Loss)}. We set $R$ to 3 in \cref{rat} to train DefectDiffu, with results shown in \cref{tab:table1} w/o AAL. It indicates that without AAL, both the quality and diversity of the generated results decrease, indicating that AAL contributes to preventing the loss of defect features and enhancing the quality of generated defect images.
\subsubsection{Multiple single-object prompts}. To validate the role of multiple single-object prompts in TDIA for improving the generated quality of defect images, we only use the text prompt from the fusion part as the condition for DefectDiffu, and the quality of generated results are shown in \cref{tab:table1} OTP. It can be seen that employing a single text prompt to control the generation of multiple objects adversely affects generation quality.
\subsubsection{Defect strength scale $w_d$.} We set $w_d$=1 and generate the same number of defect images at multiple strength scales to validate the effect of multi-strength generation on diversity. The LPIPS results are shown in \cref{tab:table1} single. It can be observed that under a single scale, the diversity is reduced, but it still outperforms the comparison methods in most products. This indicates that DefectDiffu effectively extends the diversity of the dataset through the double-free strategy.
\section{Conclusion}
This paper proposes DefectDiffu to achieve controllable generation of defect image-mask pairs across multiple products with limited samples. DefectDiffu leverages a disentangled integrated architecture and multiple single-object text prompts to disentangle defects and normal backgrounds, which facilitates the modeling of intra-product background consistency and inter-product defect consistency, thereby enhancing the diversity of normal backgrounds and defects. Furthermore, the double-free strategy is proposed to concretize the perturbation directions of the modeled background consistency and defect consistency, enabling multi-attribute controllable generation. Experimental results demonstrate that DefectDiffu outperforms state-of-the-art methods in both generation quality and diversity, effectively enhancing detection performance. Moreover, DefectDiffu achieves zero-shot defect generation based on inter-product defect consistency, reducing the costs associated with data collection and annotation.

\section*{Acknowledgements}
This work was supported by the National Key Research and Development Program of China (2022YFB3303800), the National Natural Science Foundation of China (U21A20482), and Beijing Municipal Natural Science Foundation, China, under Grant L243018.
%
%
\bibliographystyle{splncs04}
\bibliography{main}

\begin{thebibliography}{10}
\providecommand{\url}[1]{\texttt{#1}}
\providecommand{\urlprefix}{URL }
\providecommand{\doi}[1]{https://doi.org/#1}

\bibitem{Ahmed_M_R_Mathur_2020}
Ahmed, C.M., M~R, G.R., Mathur, A.P.: Challenges in machine learning based approaches for real-time anomaly detection in industrial control systems. In: Proceedings of the 6th ACM on Cyber-Physical System Security Workshop (Oct 2020). \doi{10.1145/3384941.3409588}, \url{http://dx.doi.org/10.1145/3384941.3409588}

\bibitem{Chefer_Alaluf_Vinker_Wolf_Cohen-Or_2023}
Chefer, H., Alaluf, Y., Vinker, Y., Wolf, L., Cohen-Or, D.: Attend-and-excite: Attention-based semantic guidance for text-to-image diffusion models  (Jan 2023)

\bibitem{crowson2022vqgan}
Crowson, K., Biderman, S., Kornis, D., Stander, D., Hallahan, E., Castricato, L., Raff, E.: Vqgan-clip: Open domain image generation and editing with natural language guidance. In: European Conference on Computer Vision. pp. 88--105. Springer (2022)

\bibitem{dhariwal2021diffusion}
Dhariwal, P., Nichol, A.: Diffusion models beat gans on image synthesis. Advances in neural information processing systems  \textbf{34},  8780--8794 (2021)

\bibitem{Duan_Hong_Niu_Zhang_2023}
Duan, Y., Hong, Y., Niu, L., Zhang, L.: Few-shot defect image generation via defect-aware feature manipulation  (Mar 2023)

\bibitem{feng2022training}
Feng, W., He, X., Fu, T.J., Jampani, V., Akula, A., Narayana, P., Basu, S., Wang, X.E., Wang, W.Y.: Training-free structured diffusion guidance for compositional text-to-image synthesis. arXiv preprint arXiv:2212.05032  (2022)

\bibitem{gal2022image}
Gal, R., Alaluf, Y., Atzmon, Y., Patashnik, O., Bermano, A.H., Chechik, G., Cohen-Or, D.: An image is worth one word: Personalizing text-to-image generation using textual inversion. arXiv preprint arXiv:2208.01618  (2022)

\bibitem{10.5555/2969033.2969125}
Goodfellow, I.J., Pouget-Abadie, J., Mirza, M., Xu, B., Warde-Farley, D., Ozair, S., Courville, A., Bengio, Y.: Generative adversarial nets. In: Proceedings of the 27th International Conference on Neural Information Processing Systems - Volume 2. p. 2672–2680. NIPS'14, MIT Press, Cambridge, MA, USA (2014)

\bibitem{Ho_Jain_Abbeel_2020}
Ho, J., Jain, A., Abbeel, P.: Denoising diffusion probabilistic models. Neural Information Processing Systems,Neural Information Processing Systems  (Jan 2020)

\bibitem{ho2022classifier}
Ho, J., Salimans, T.: Classifier-free diffusion guidance. arXiv preprint arXiv:2207.12598  (2022)

\bibitem{Hu_Zhang_Yi_Du_Chen_Liu_Wang_Wang_2023}
Hu, T., Zhang, J., Yi, R., Du, Y., Chen, X., Liu, L., Wang, Y., Wang, C.: Anomalydiffusion: Few-shot anomaly image generation with diffusion model  (Dec 2023)

\bibitem{Jeong_Zou_Kim_Zhang_Ravichandran_Dabeer_2023}
Jeong, J., Zou, Y., Kim, T., Zhang, D., Ravichandran, A., Dabeer, O.: Winclip: Zero-/few-shot anomaly classification and segmentation  (Mar 2023)

\bibitem{9156570}
Karras, T., Laine, S., Aittala, M., Hellsten, J., Lehtinen, J., Aila, T.: Analyzing and improving the image quality of stylegan. In: 2020 IEEE/CVF Conference on Computer Vision and Pattern Recognition (CVPR). pp. 8107--8116 (2020). \doi{10.1109/CVPR42600.2020.00813}

\bibitem{Kingma2014}
Kingma, D.P., Welling, M.: {Auto-Encoding Variational Bayes}. In: 2nd International Conference on Learning Representations, {ICLR} 2014, Banff, AB, Canada, April 14-16, 2014, Conference Track Proceedings (2014)

\bibitem{KowalskiECCV2020}
Kowalski, M., Garbin, S.J., Estellers, V., Baltrušaitis, T., Johnson, M., Shotton, J.: Config: Controllable neural face image generation. In: European Conference on Computer Vision (ECCV) (2020)

\bibitem{kulikov2023sinddm}
Kulikov, V., Yadin, S., Kleiner, M., Michaeli, T.: Sinddm: A single image denoising diffusion model. In: International Conference on Machine Learning. pp. 17920--17930. PMLR (2023)

\bibitem{kumari2023multi}
Kumari, N., Zhang, B., Zhang, R., Shechtman, E., Zhu, J.Y.: Multi-concept customization of text-to-image diffusion. In: Proceedings of the IEEE/CVF Conference on Computer Vision and Pattern Recognition. pp. 1931--1941 (2023)

\bibitem{Liu_Liu_Zheng_Wang_Mao_Qiu_Ling_2023}
Liu, R., Liu, W., Zheng, Z., Wang, L., Mao, L., Qiu, Q., Ling, G.: Anomaly-gan: A data augmentation method for train surface anomaly detection. Expert Systems with Applications p. 120284 (Oct 2023). \doi{10.1016/j.eswa.2023.120284}, \url{http://dx.doi.org/10.1016/j.eswa.2023.120284}

\bibitem{Niu_Li_Wang_Lin_2020}
Niu, S., Li, B., Wang, X., Lin, H.: Defect image sample generation with gan for improving defect recognition. IEEE Transactions on Automation Science and Engineering p. 1–12 (Jan 2020). \doi{10.1109/tase.2020.2967415}, \url{http://dx.doi.org/10.1109/tase.2020.2967415}

\bibitem{Niu_Li_Wang_Peng_2022}
Niu, S., Li, B., Wang, X., Peng, Y.: Region- and strength-controllable gan for defect generation and segmentation in industrial images. IEEE Transactions on Industrial Informatics p. 4531–4541 (Jul 2022). \doi{10.1109/tii.2021.3127188}, \url{http://dx.doi.org/10.1109/tii.2021.3127188}

\bibitem{Niu_Peng_Li_Wang}
Niu, S., Peng, Y., Li, B., Wang, X.: A transformed-feature-space data augmentation method for defect segmentation

\bibitem{peebles2023scalable}
Peebles, W., Xie, S.: Scalable diffusion models with transformers. In: Proceedings of the IEEE/CVF International Conference on Computer Vision. pp. 4195--4205 (2023)

\bibitem{radford2021learning}
Radford, A., Kim, J.W., Hallacy, C., Ramesh, A., Goh, G., Agarwal, S., Sastry, G., Askell, A., Mishkin, P., Clark, J., et~al.: Learning transferable visual models from natural language supervision. In: International conference on machine learning. pp. 8748--8763. PMLR (2021)

\bibitem{Ramesh_Dhariwal_Nichol_Chu_Chen}
Ramesh, A., Dhariwal, P., Nichol, A., Chu, C., Chen, M.: Hierarchical text-conditional image generation with clip latents

\bibitem{Rombach_Blattmann_Lorenz_Esser_Ommer_2022}
Rombach, R., Blattmann, A., Lorenz, D., Esser, P., Ommer, B.: High-resolution image synthesis with latent diffusion models. In: 2022 IEEE/CVF Conference on Computer Vision and Pattern Recognition (CVPR) (Jun 2022). \doi{10.1109/cvpr52688.2022.01042}, \url{http://dx.doi.org/10.1109/cvpr52688.2022.01042}

\bibitem{Ruiz_Li_Jampani_Pritch_Rubinstein_Aberman_2022}
Ruiz, N., Li, Y., Jampani, V., Pritch, Y., Rubinstein, M., Aberman, K.: Dreambooth: Fine tuning text-to-image diffusion models for subject-driven generation  (Aug 2022)

\bibitem{schluter2022natural}
Schl{\"u}ter, H.M., Tan, J., Hou, B., Kainz, B.: Natural synthetic anomalies for self-supervised anomaly detection and localization. In: European Conference on Computer Vision. pp. 474--489. Springer (2022)

\bibitem{Singh_Desai_2023}
Singh, S.A., Desai, K.A.: Automated surface defect detection framework using machine vision and convolutional neural networks. Journal of Intelligent Manufacturing p. 1995–2011 (Apr 2023). \doi{10.1007/s10845-021-01878-w}, \url{http://dx.doi.org/10.1007/s10845-021-01878-w}

\bibitem{tang2020xinggan}
Tang, H., Bai, S., Zhang, L., Torr, P.H., Sebe, N.: Xinggan for person image generation. In: ECCV (2020)

\bibitem{machines10121239}
Wei, J., Zhang, Z., Shen, F., Lv, C.: Mask-guided generation method for industrial defect images with non-uniform structures. Machines  \textbf{10}(12) (2022). \doi{10.3390/machines10121239}, \url{https://www.mdpi.com/2075-1702/10/12/1239}

\bibitem{Wu_Zhao_Shou_Zhou_Shen_2023}
Wu, W., Zhao, Y., Shou, M., Zhou, H., Shen, C.: Diffumask: Synthesizing images with pixel-level annotations for semantic segmentation using diffusion models  (Mar 2023)

\bibitem{yang2023defect}
Yang, S., Chen, Z., Chen, P., Fang, X., Liu, S., Chen, Y.: Defect spectrum: A granular look of large-scale defect datasets with rich semantics. arXiv preprint arXiv:2310.17316  (2023)

\bibitem{Yoshihashi_Otsuka_Doi_Tanaka_2023}
Yoshihashi, R., Otsuka, Y., Doi, K., Tanaka, T.: Attention as annotation: Generating images and pseudo-masks for weakly supervised semantic segmentation with diffusion  (Sep 2023)

\bibitem{Zavrtanik_Kristan_Skočaj}
Zavrtanik, V., Kristan, M., Skočaj, D.: Dsr -a dual subspace re-projection network for surface anomaly detection

\bibitem{Zhang_Cui_Hung_Lu_2021}
Zhang, G., Cui, K., Hung, T.Y., Lu, S.: Defect-gan: High-fidelity defect synthesis for automated defect inspection. In: 2021 IEEE Winter Conference on Applications of Computer Vision (WACV) (Jan 2021). \doi{10.1109/wacv48630.2021.00257}, \url{http://dx.doi.org/10.1109/wacv48630.2021.00257}

\bibitem{zhang2021defect}
Zhang, G., Cui, K., Hung, T.Y., Lu, S.: Defect-gan: High-fidelity defect synthesis for automated defect inspection. In: Proceedings of the IEEE/CVF Winter Conference on Applications of Computer Vision. pp. 2524--2534 (2021)

\bibitem{Zhang_Agrawala}
Zhang, L., Agrawala, M.: Adding conditional control to text-to-image diffusion models

\bibitem{zhao2023uni}
Zhao, S., Chen, D., Chen, Y.C., Bao, J., Hao, S., Yuan, L., Wong, K.Y.K.: Uni-controlnet: All-in-one control to text-to-image diffusion models. Advances in Neural Information Processing Systems  (2023)

\bibitem{Zheng_2023_CVPR}
Zheng, G., Zhou, X., Li, X., Qi, Z., Shan, Y., Li, X.: Layoutdiffusion: Controllable diffusion model for layout-to-image generation. In: Proceedings of the IEEE/CVF Conference on Computer Vision and Pattern Recognition (CVPR). pp. 22490--22499 (June 2023)

\bibitem{Zhou_Pang_Tian_He_Chen_2023}
Zhou, Q., Pang, G., Tian, Y., He, S., Chen, J.: Anomalyclip: Object-agnostic prompt learning for zero-shot anomaly detection  (Oct 2023)

\bibitem{zhou2023maskdiffusion}
Zhou, Y., Zhou, D., Zhu, Z.L., Wang, Y., Hou, Q., Feng, J.: Maskdiffusion: Boosting text-to-image consistency with conditional mask. arXiv preprint arXiv:2309.04399  (2023)

\end{thebibliography}


\section*{Supplementary material}

# Few-shot Defect Image Generation based on Consistency Modeling

Supplementary Material

## 1 Supplementary Experiment

### 1.1 Implementation

To model the inter-product defect consistency direction, we modify the defect names of similar defects in the MVTecAD dataset. The defect names before and after modification are shown in Tab. 2.

### 1.2 Implementation for Comparison Methods

**StyleGAN2:** We conduct training for 3000 iterations on StyleGAN2 for each defect category, with a batch size of 4. Other parameters remained consistent with the specifications outlined in the official documentation.

**DFMGAN:** Firstly, StyleGAN2 is trained on defect-free images of each product for 3000 iterations using a batch size of 4. Subsequently, the pre-trained StyleGAN2 is finetuned to defect images of each defect category, employing a batch size of 4. All other parameters adhered to the official documentation.

**Defect-Gen:** Both models undergo training for 200,000 iterations, with all other settings aligned with the official documentation.

### 1.3 Comparison of Classification

To verify the role of generated defect images in classification tasks, we employ the same dataset as in the segmentation experiments to train the ResNet50 for binary classification. Each product is trained for 2000 epochs with a learning rate of 0.0002 and a batch size of 12. The average classification accuracy on MVTec is shown in Tab. 1, where our DefectDiffu achieves the optimal classification performance.

**Table 1:** Comparison of Classification Experiments. AA represents Average Accuracy.

| Method | Baseline | DefectDiffu(Ours) | DFMGAN | Defect-Gen | StyleGAN |
|--------|----------|-------------------|--------|------------|----------|
| AA     | 81.13%   | 91.74%            | 88.15% | 89.75%     | 87.26%   |

**Table 2:** MVTecAD Dataset Defect Names: *defect* refers to the original defect name, *uni_defect* refers to the modified defect name used in training, and *product* refers to the product name

| product | defect | uni_defect | product | defect | uni_defect |
|---|---|---|---|---|---|
| metal_nut | scratch | scratch | hazelnut | print | print |
|  | flip | flip |  | hole | hole |
|  | color | color |  | cut | cut |
|  | bent | bent |  | crack | crack |
| carpet | hole | hole | wood | scratch | scratch |
|  | metal_contamination | metal_contamination |  | combined | combined |
|  | cut | cut |  | hole | hole |
|  | thread | thread |  | liquid | liquid |
|  | color | color |  | color | color |
| screw | scratch_neck | scratch | capsule | scratch | scratch |
|  | thread_top | thread |  | faulty_imprint | faulty_imprint |
|  | scratch_head | scratch |  | squeeze | squeeze |
|  | manipulated_front | manipulated_front |  | crack | crack |
|  | thread_side | thread |  | poke | hole |
| leather | glue | liquid | grid | broken | broken |
|  | cut | cut |  | glue | liquid |
|  | color | color |  | metal_contamination | metal_contamination |
|  | poke | hole |  | thread | thread |
|  | fold | fold |  | bent | bent |
| pill | scratch | scratch | zipper | rough | rough |
|  | faulty_imprint | faulty_imprint |  | fabric_border | cut |
|  | combined | combined |  | combined | combined |
|  | contamination | contamination |  | fabric_interior | cut |
|  | pill_type | pill_type |  | squeezed_teeth | squeeze |
|  | crack | crack |  | broken_teeth | broken |
|  | color | color |  | split_teeth | split |
| cable | missing_cable | missing | tile | rough | rough |
|  | missing_wire | missing |  | gray_stroke | color |
|  | cable_swap | cable_swap |  | glue_strip | glue_strip |
|  | bent_wire | bent |  | oil | liquid |
|  | poke_insulation | hole |  | crack | crack |
|  | combined | combined | bottle | contamination | contamination |
|  | cut_inner_insulation | cut |  | broken_small | broken |
|  | cut_outer_insulation | cut |  | broken_large | broken |
| transistor | cut_lead | cut | toothbrush | defective | bent |
|  | damaged_case | damaged_case |  |  |  |
|  | misplaced | misplaced |  |  |  |
|  | bent_lead | bent |  |  |  |

### 1.4 Experiments on the VISA and KSDD datasets

We tested our method on the KSDD and VISA datasets under the same settings. The experimental results are presented in Table A and Table B.

**Table A:** Comparison of average quantitative results.

| Dataset | Index | Baseline | DefectDiffu | DefectGen | DFMGAN |
|---|---|---|---|---|---|
| VisA | FID | \ | **39.24** | 185.91 | 180.15 |
| | LPIPS (%) | \ | **10.84** | 10.57 | 7.15 |
| | IoU (%) | 44.44 | **53.34** | 47.88 | 46.78 |
| | F1 (%) | 57.75 | **66.65** | 61.72 | 60.49 |
| | AP (%) | 63.96 | **71.26** | 68.15 | 67.19 |
| | AUC (%) | 98.16 | **98.81** | 98.25 | 98.15 |
| KSDD2 | FID | \ | **47.92** | 320.60 | 365.56 |
| | LPIPS (%) | \ | 2.89 | **4.32** | 1.75 |
| | IoU (%) | 63.96 | **66.08** | 65.17 | 63.32 |
| | F1 (%) | 78.02 | **79.58** | 78.91 | 77.54 |
| | AP (%) | 85.33 | **87.52** | 87.32 | 85.50 |
| | AUC (%) | 98.66 | **99.53** | 99.45 | 99.36 |

**Table B:** Comparison of mIoU across each product.

| VisA | candle | capsules | cashew | chewinggum | fryum | macaroni1 | macaroni2 | pcb1 | pcb2 | pcb3 | pcb4 | pipe_fryum |
|---|---|---|---|---|---|---|---|---|---|---|---|---|
| Baseline | 8.06 | 45.58 | 81.32 | 54.01 | 69.12 | 25.49 | 15.52 | 66.46 | 27.26 | 26.43 | 39.03 | 75.04 |
| DefectDiffu | **29.18** | **62.12** | **86.89** | **61.81** | **79.83** | **28.08** | 24.19 | 70.89 | **32.51** | **27.22** | **49.84** | **87.57** |
| DefectGen | 24.90 | 60.64 | 82.08 | 59.85 | 62.85 | 24.04 | 23.23 | 65.13 | 32.33 | 18.62 | 42.10 | 78.83 |
| DFMGAN | 19.95 | 49.54 | 79.84 | 57.34 | 78.88 | 27.56 | 22.04 | **71.11** | 23.95 | 19.28 | 49.64 | 62.21 |

### 1.5 Additional Zero-shot Generation Results

We validate the effectiveness of zero-shot generation on wood, tile, and leather, and the results are depicted in Fig. 1.

### 1.6 Controllable Generation Results for All Types of Defects Strengths

We perform controllable defect image generation for all defect products in the MVTecAD dataset. The generated results for various defect types are shown in Fig. 2-Fig. 16, demonstrating the strong controllability of our model over defect strengths.

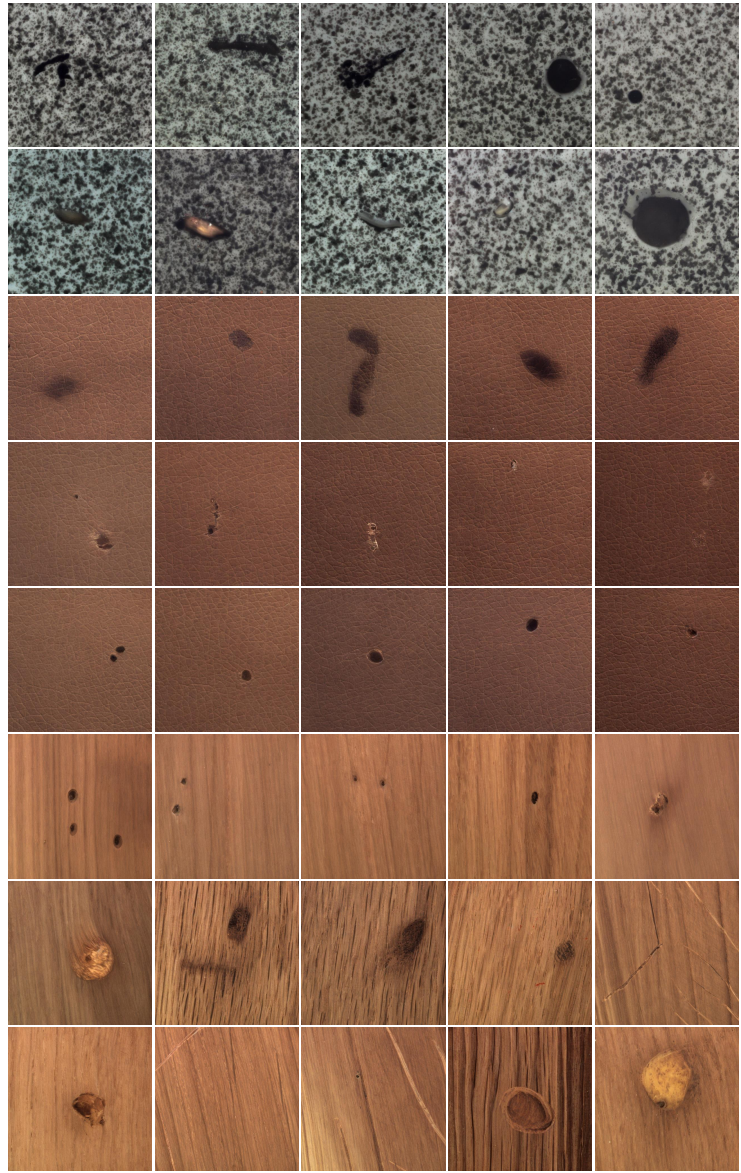

**Fig. 1:** Zero-shot defect generation results of tile, leather, and wood.

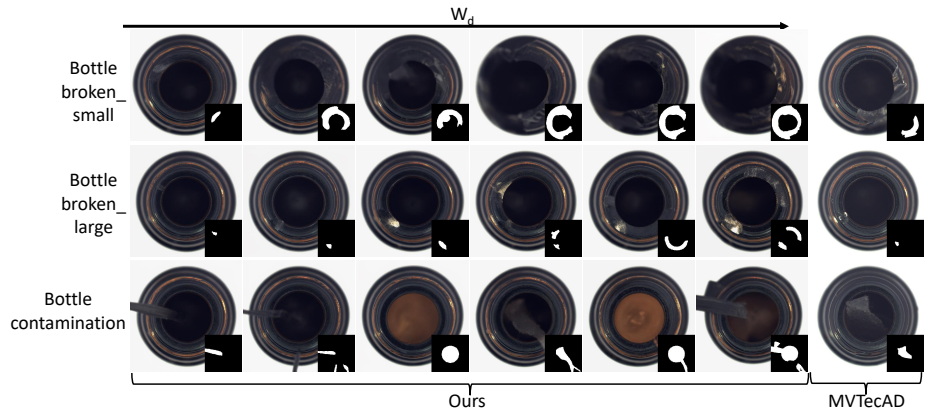

**Fig. 2:** Results of varying defect strengths for different types of defects under the category bottle. *MVTecAD* represents real data, while $w_d$ represents increasing defect strengths from left to right.

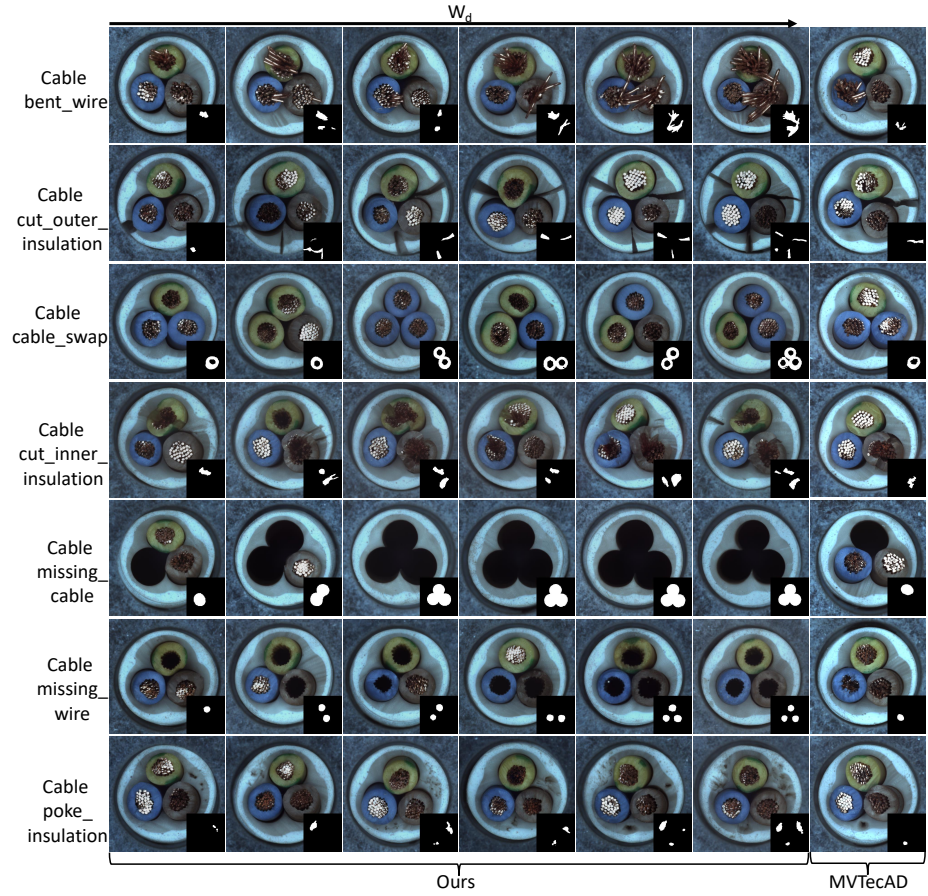

**Fig. 3:** Result of varying defect strengths for different types of defects under the category cable. *MVTecAD* represents real data, while $w_d$ represents increasing defect strengths from left to right.

Abbreviated paper title 7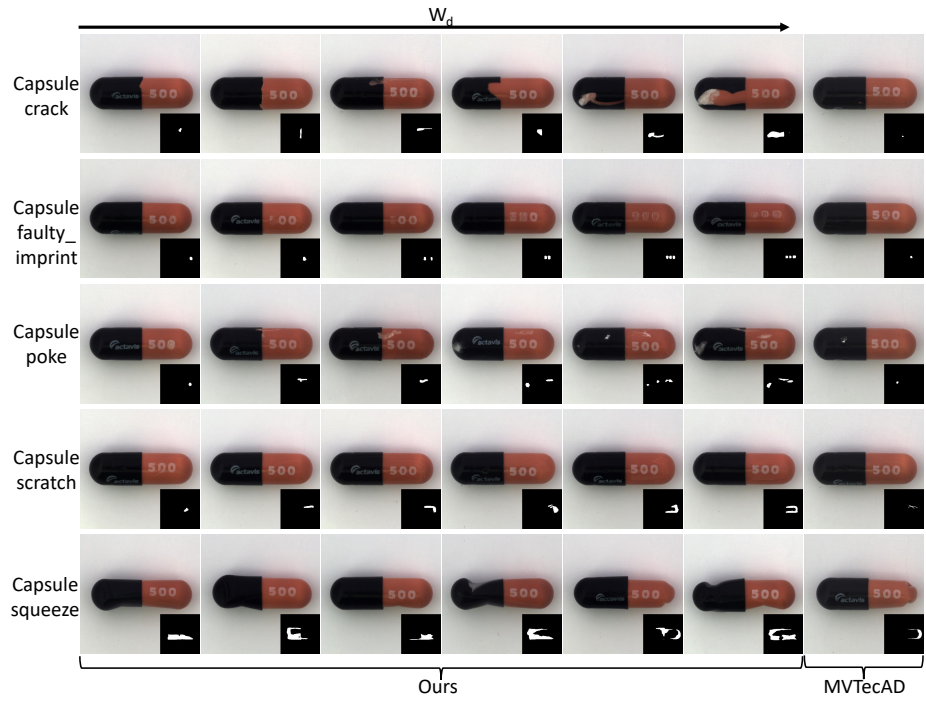

**Fig. 4:** Result of varying defect strengths for different types of defects under the category capsule. *MVTecAD* represents real data, while $w_d$ represents increasing defect strengths from left to right.

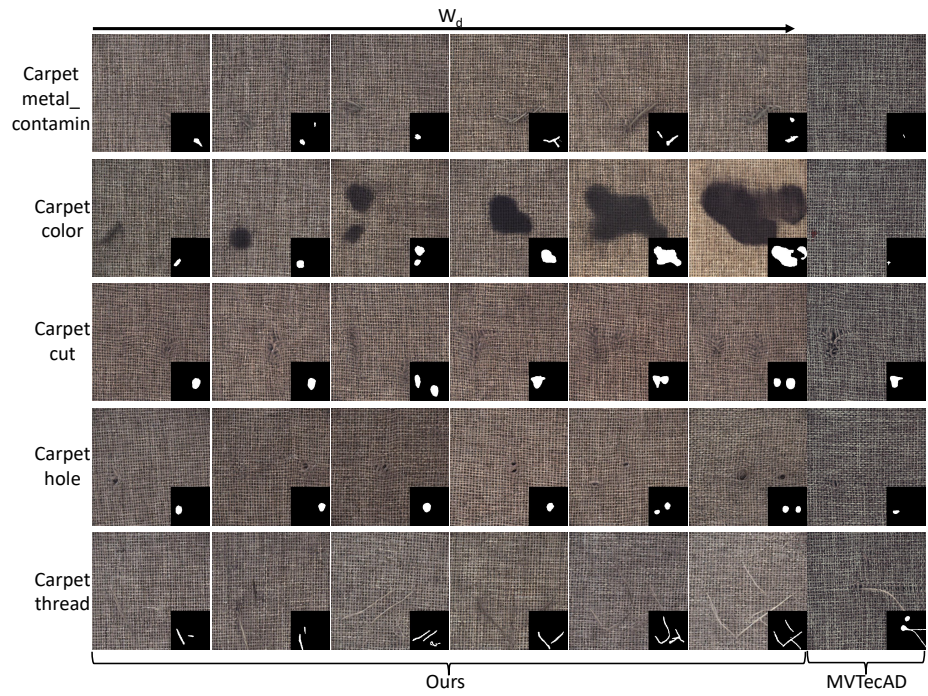

**Fig. 5:** Result of varying defect strengths for different types of defects under the category carpet. *MVTecAD* represents real data, while $w_d$ represents increasing defect strengths from left to right.

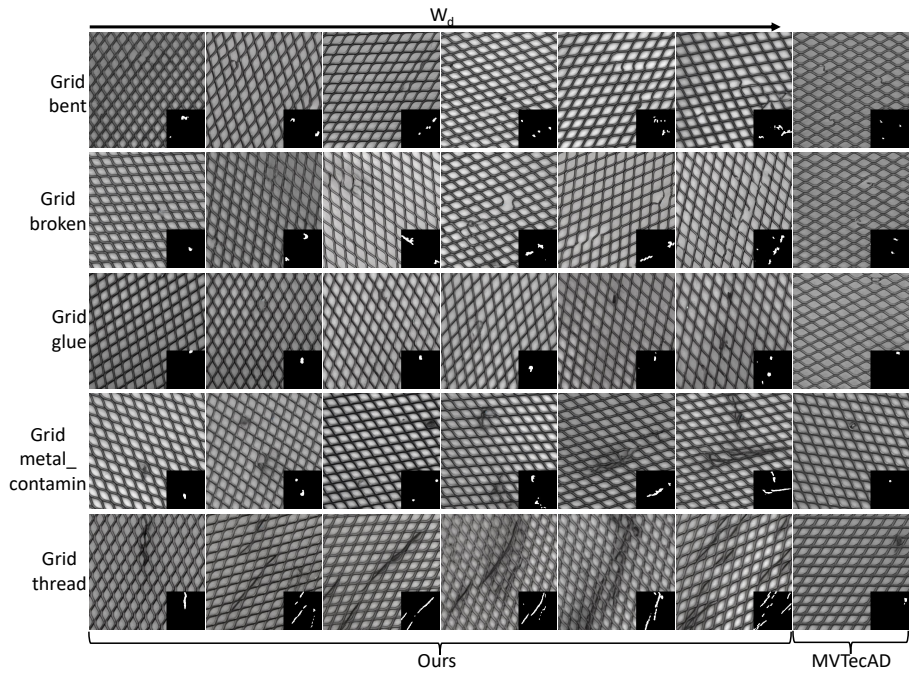

**Fig. 6:** Result of varying defect strengths for different types of defects under the category grid. *MVTecAD* represents real data, while $w_d$ represents increasing defect strengths from left to right.

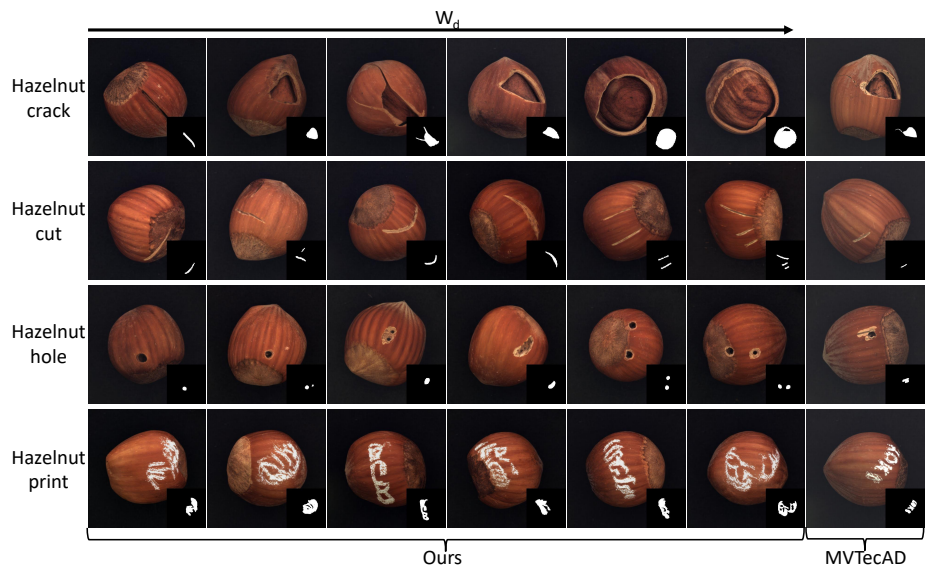

**Fig. 7:** Result of varying defect strengths for different types of defects under the category hazelnut. *MVTecAD* represents real data, while $w_d$ represents increasing defect strengths from left to right.

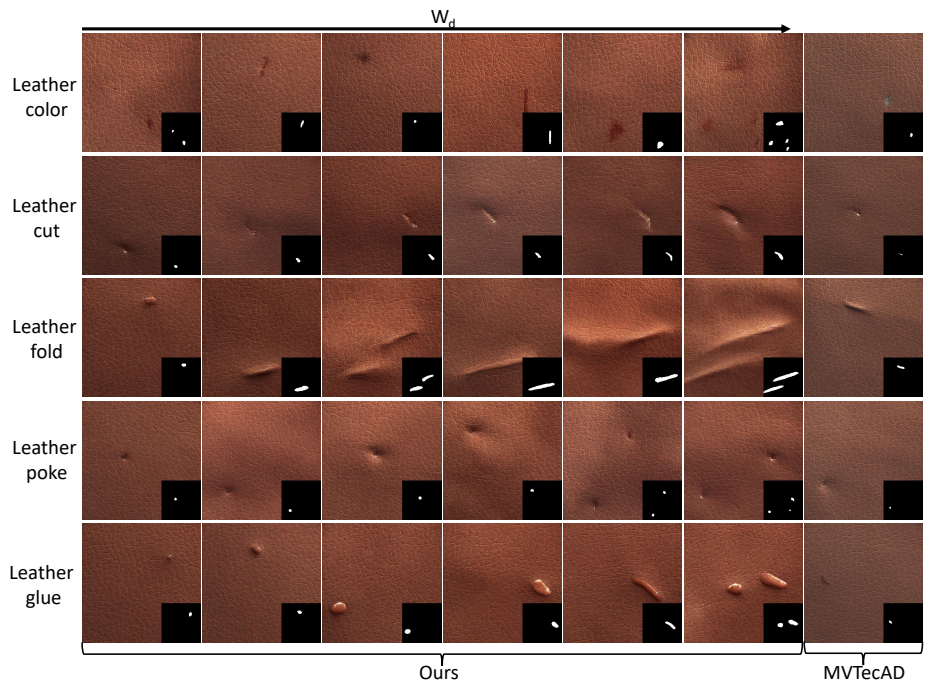

**Fig. 8:** Result of varying defect strengths for different types of defects under the category leather. *MVTecAD* represents real data, while $w_d$ represents increasing defect strengths from left to right.

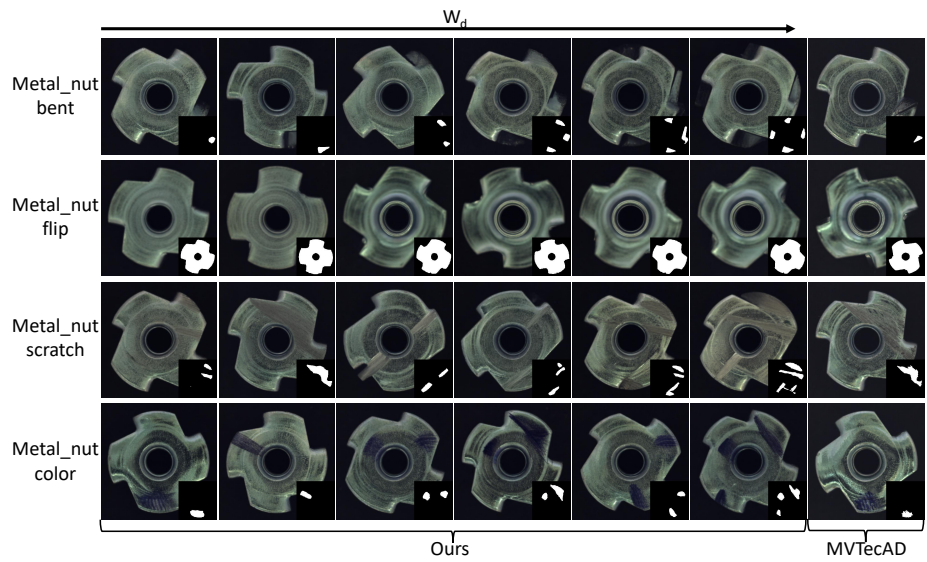

**Fig. 9:** Result of varying defect strengths for different types of defects under the category metal_nut. *MVTecAD* represents real data, while $w_d$ represents increasing defect strengths from left to right.

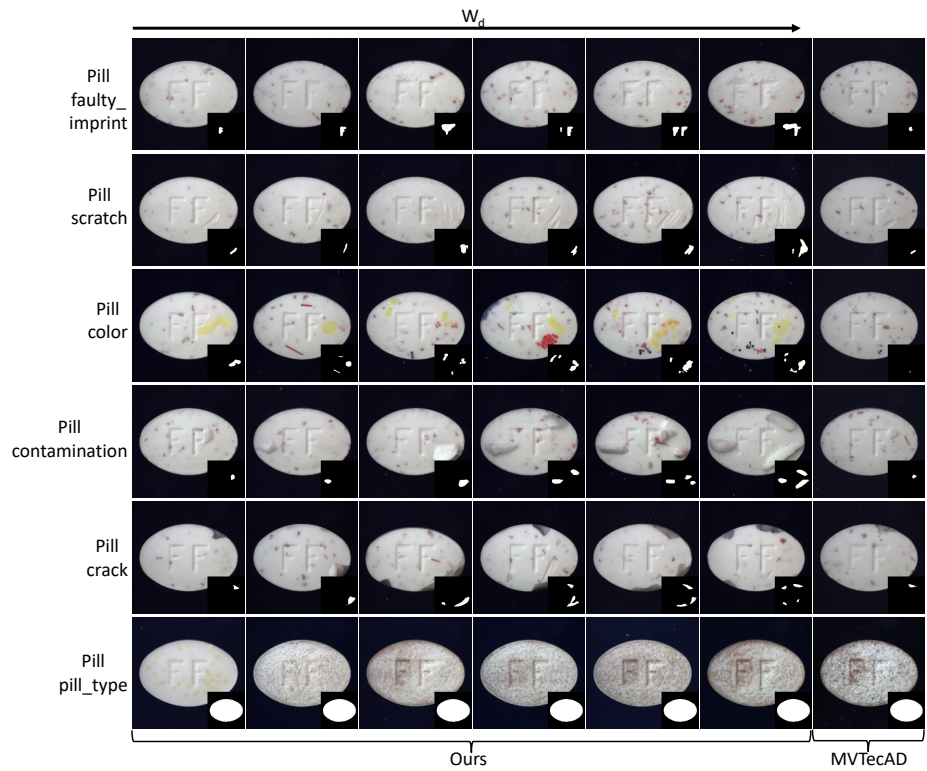

**Fig. 10:** Result of varying defect strengths for different types of defects under the category pill. *MVTecAD* represents real data, while $w_d$ represents increasing defect strengths from left to right.

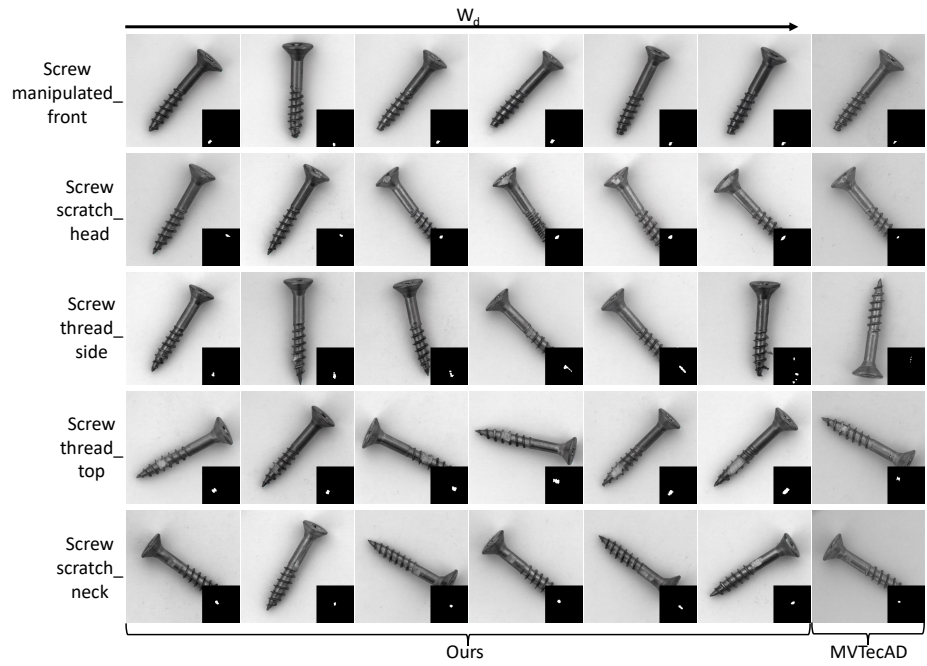

Fig. 11: Result of varying defect strengths for different types of defects under the category screw. *MVTecAD* represents real data, while $w_d$ represents increasing defect strengths from left to right.

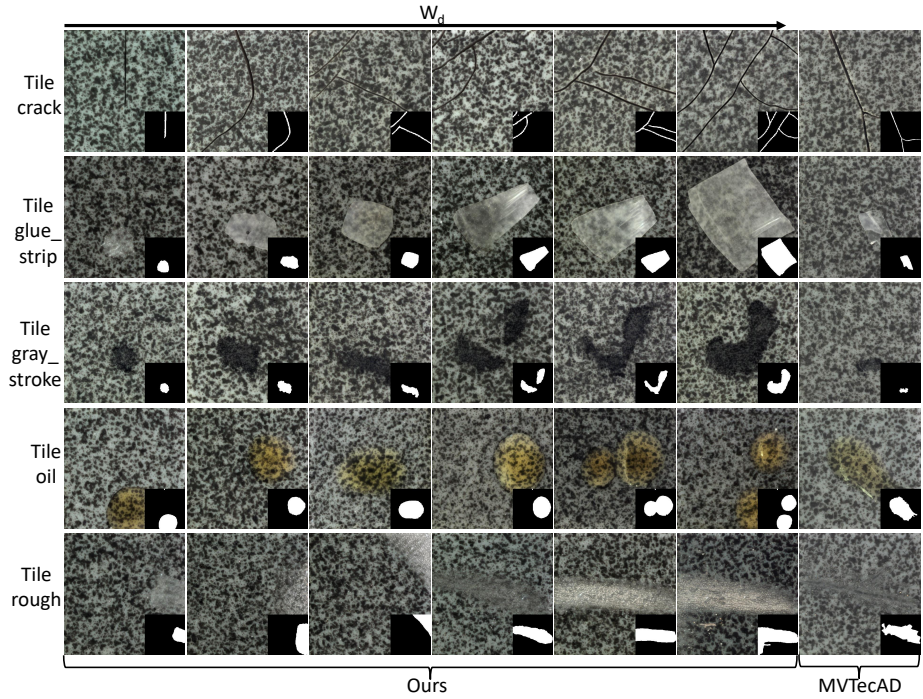

**Fig. 12:** Result of varying defect strengths for different types of defects under the category tile. *MVTecAD* represents real data, while $w_d$ represents increasing defect strengths from left to right.

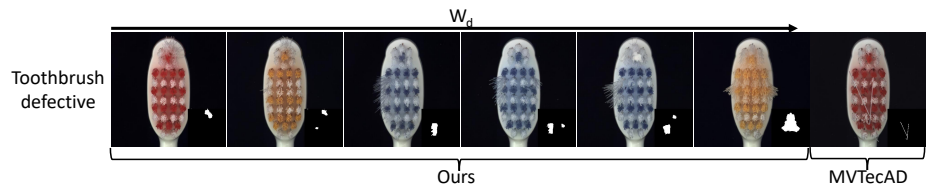

**Fig. 13:** Result of varying defect strengths for different types of defects under the category toothbrush. *MVTecAD* represents real data, while $w_d$ represents increasing defect strengths from left to right.

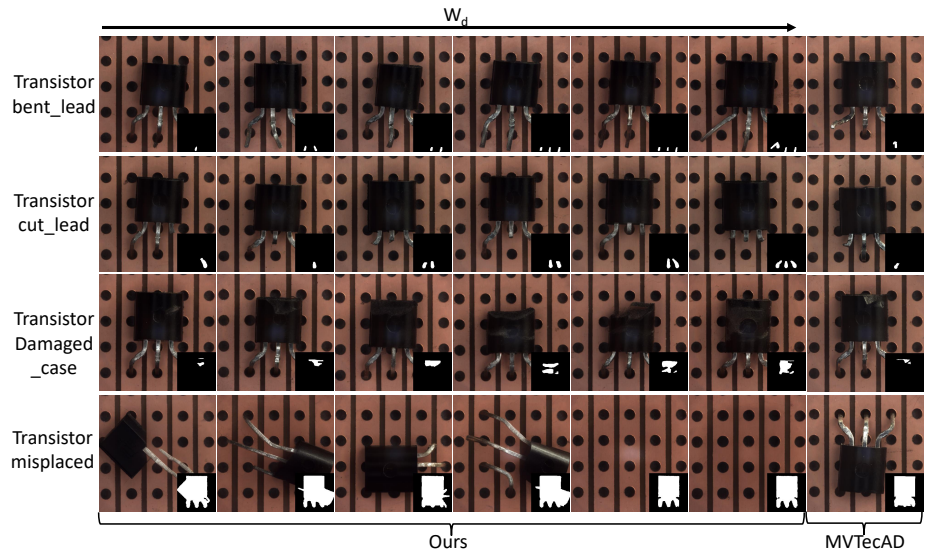

**Fig. 14:** Result of varying defect strengths for different types of defects under the category transistor. *MVTecAD* represents real data, while $w_d$ represents increasing defect strengths from left to right.

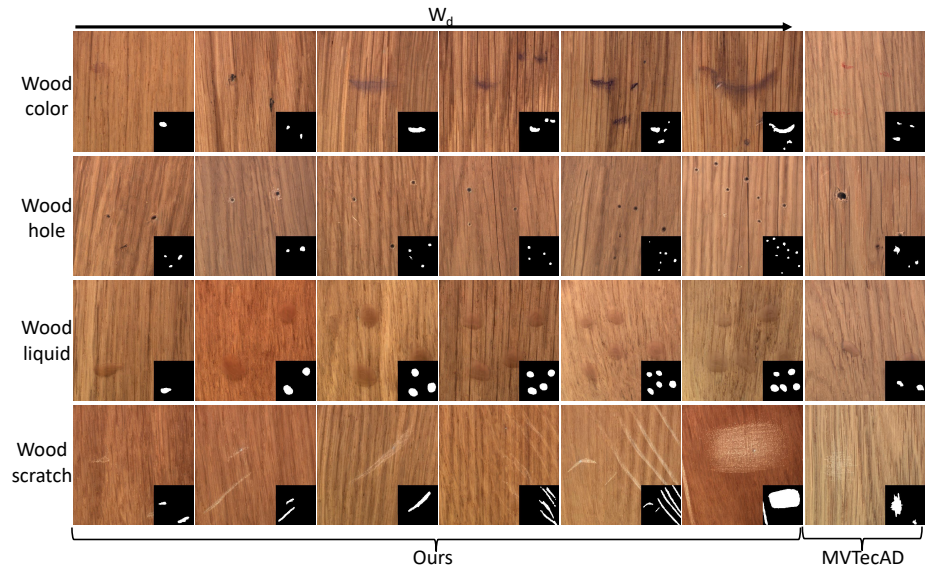

**Fig. 15:** Result of varying defect strengths for different types of defects under the category wood. *MVTecAD* represents real data, while $w_d$ represents increasing defect strengths from left to right.

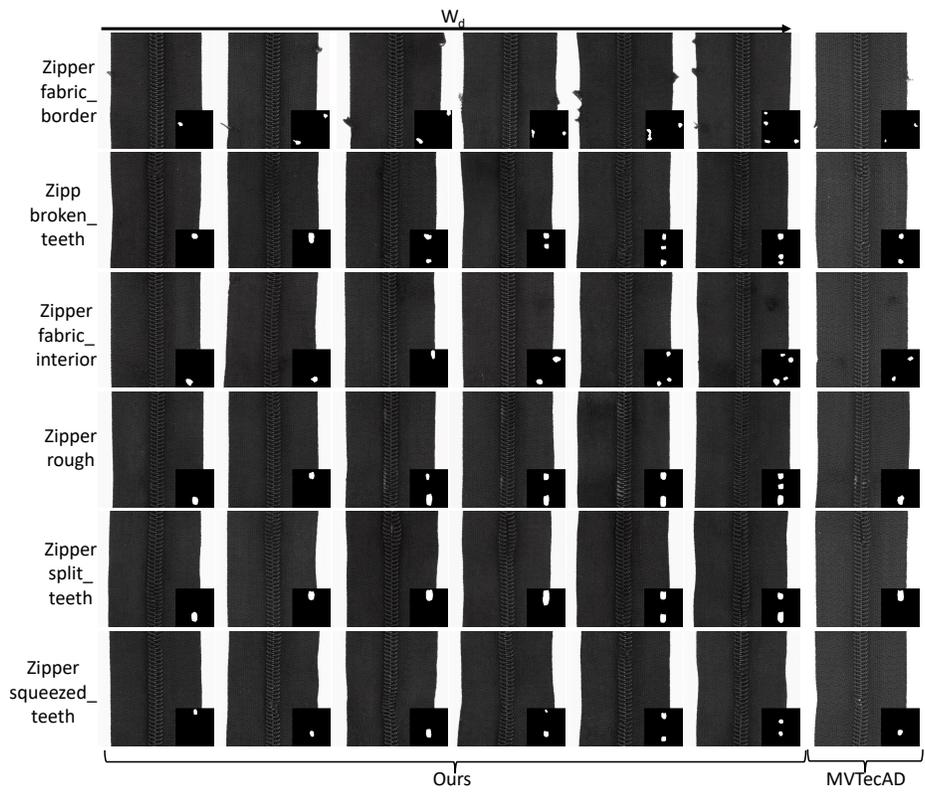

**Fig. 16:** Result of varying defect strengths for different types of defects under the category zipper. *MVTecAD* represents real data, while $w_d$ represents increasing defect strengths from left to right.


\end{document}